\documentclass[letterpaper]{article} 
\usepackage{aaai23}  
\usepackage{times}  
\usepackage{helvet}  
\usepackage{courier}  
\usepackage[hyphens]{url}  
\usepackage{graphicx} 
\usepackage{natbib}  
\usepackage{caption} 
\usepackage{ifpdf}
\ifpdf
\else
  
\fi
\usepackage{algorithm}
\usepackage{algorithmic}
\usepackage{xcolor}
\usepackage{subcaption}

\usepackage{booktabs}
\usepackage{amsmath}
\usepackage{placeins}

\usepackage{pifont}
\newcommand{\xmark}{\ding{55}}
\providecommand{\checkmark}{\ding{51}}

\usepackage{tikz}
\usetikzlibrary{arrows.meta,positioning,shapes.geometric,fit,backgrounds,calc}

\usepackage{newfloat}
\usepackage{listings}
\DeclareCaptionStyle{ruled}{labelfont=normalfont,labelsep=colon,strut=off} 
\floatstyle{ruled}
\newfloat{listing}{tb}{lst}{}
\floatname{listing}{Listing}
\ifpdf
\fi

\title{Towards Inclusive Toxic Content Moderation: Addressing Vulnerabilities to Adversarial Attacks in Toxicity Classifiers Tackling LLM-generated Content}
\author{
Shaz Furniturewala\textsuperscript{1},
Arkaitz Zubiaga\textsuperscript{2}
}
\affiliations{
\textsuperscript{1}Center for Data Science, New York University,
\textsuperscript{2}Queen Mary University of London \\
\texttt{mohammadshaz529@gmail.com}
}

\begin{document}

\maketitle

\begin{abstract}
Adversarial perturbations can reduce state-of-the-art toxicity classifiers to near-zero accuracy, yet existing defences treat models as black boxes. We apply mechanistic interpretability to toxicity classification for the first time, identifying the internal attention-head circuits responsible for both correct classification and adversarial vulnerability. Across a 2$\times$2 factorial study (BERT $\times$ RoBERTa) $\times$ (Jigsaw $\times$ ToxiGen), extended to Llama Guard~2 (8B), we show that zeroing a single attention head recovers up to \textbf{70.4 pp} of adversarial accuracy for RoBERTa on Jigsaw and \textbf{37.3 pp} for Llama Guard~2 on ToxiGen, at $\leq$0.6 pp clean cost. Vulnerable heads generalise to held-out examples within $\leq$1 pp, and a class-imbalance sweep confirms they act as selective toxic-class detectors. Head suppression matches or outperforms adversarial training on Jigsaw; data augmentation dominates on ToxiGen: a dataset-specific reversal explained by whether the classifier encodes a concentrated bottleneck or a distributed circuit. Demographic analysis across 20 Jigsaw and 13 ToxiGen minority groups reveals structurally unequal adversarial vulnerability, exposing mechanistically traceable fairness gaps in current toxicity classifiers.
\end{abstract}

\begin{table*}[t]
\centering
\begin{tabular}{lp{0.52\linewidth}p{0.1\linewidth}p{0.16\linewidth}}
\toprule
\textbf{Input} & \textbf{Text} & \textbf{Label} & \textbf{Prediction} \\
\midrule
\textbf{Regular} & ``Your observations are consistent with those of most imbeciles'' & Toxic & Toxic \checkmark \\
\textbf{Adversarial} & ``your terms are consistent with those of most imbeciles'' & Toxic & Non-toxic \xmark \\ \midrule
\textbf{Regular} & ``Homosexuality is a sin and a choice, not a genetic factor. Homosexuality is a behavioral disorder conflicting with one's God-given sex'' & Toxic & Toxic \checkmark \\
\textbf{Adversarial} & ``homosexuality is a sin and a choice, not a genetic factor. homosexual is a behavioral disorder conflicting with one ' s god - given sex'' & Toxic & Non-toxic \xmark \\ \midrule
\textbf{Regular} & ``the immigration system in the USA is full of absurdities'' & Non-toxic & Non-toxic \checkmark \\
\textbf{Adversarial} & ``the immigration system in the us is full of absurdities'' & Non-toxic & Toxic \xmark \\
\bottomrule
\end{tabular}
\caption{Adversarial input example showing semantic equivalence but different model predictions.}
\label{tab:adv_example_transposed}
\end{table*}

\section{\textbf{Introduction}}
With the advent of Large Language Models (LLMs), the Internet is seeing more machine-generated content than ever before \cite{wu-etal-2025-survey, brooks-etal-2024-rise}, including toxic content \cite{BARMAN2024100545}. While content moderation systems have previously struggled to accurately classify this content \cite{Warner_2025}, the ability of users to automatically generate such toxic content using LLMs introduces a new challenge that is currently understudied. Most toxicity classification models are trained on human-generated text and on labeled data from a specific source, for example, Twitter posts \cite{mollas_ethos_2022, casellietal2020}. This distributional mismatch means classifiers trained on human text frequently misclassify LLM-generated toxicity, and adversarially perturbed text can evade detection entirely \cite{yin2021towards, toxigen}. Research into adversarial prompting shows that it is possible to reduce the accuracy of text-transformer classifiers to nearly zero with an effectively imperceptible change in the input \cite{guo2021gradientbasedadversarialattackstext, tokenmodification, Sadrizadeh2023ACA, Li_2019}.

Existing defenses treat models as black boxes, relying on external detectors \cite{chen2022should, yoo-etal-2022-detection} or adversarial training against known attacks \cite{bespalov2024buildingrobusttoxicitypredictor, samory_sexism} -- strategies that require continuous retraining and remain reactive rather than proactive.

In this work, we leverage mechanistic interpretability to move beyond black-box defenses. Circuits -- interpretable subnetworks corresponding to specific model behaviours \cite{Olah2020} -- have been studied in toy classification settings \cite{Garc_a_Carrasco_2024}; we extend this framework to toxicity classification. By zero-ablating each attention head, we identify which heads are crucial for correct classification and which are exploited by adversarial attacks. A further open question is \emph{who} suffers when classifiers fail: prior work on demographic bias has focused on clean-input disparities \cite{dixon2018measuring}, but has not examined whether adversarial vulnerability is itself unequally distributed across minority groups.

By conducting for the first time a comprehensive mechanistic interpretability study on toxic content moderation, our work makes the following novel contributions:
\begin{itemize}\setlength{\itemsep}{1pt}\setlength{\parsep}{0pt}
 \item We conduct a \textbf{full 2$\times$2 factorial study} (BERT, RoBERTa) $\times$ (Jigsaw, ToxiGen) covering all phases from PGD adversarial generation through clean/adversarial attention patching, demographic analysis, robustness interventions, split-sample validation, and a controlled class-imbalance sweep, and extend the entire pipeline to Llama Guard~2 (8B).
 \item We follow a mechanistic interpretability workflow \cite{conmy2023automatedcircuitdiscoverymechanistic, Garc_a_Carrasco_2024}, to date unstudied in the context of toxicity detection, to identify heads that are \textit{crucial} to clean classification and heads that are \textit{vulnerable} to adversarial attack. We show that a single head accounts for up to 70.4 pp of adversarial accuracy recovery for RoBERTa on Jigsaw, and 37.3 pp for Llama Guard~2 on ToxiGen.
 \item We demonstrate that head suppression is \textit{dataset-specific}: it matches or exceeds adversarial training on Jigsaw (the single-bottleneck regime) but is outperformed by data augmentation on ToxiGen (the distributed-circuit regime).
 \item We provide demographic-level analysis across 20 Jigsaw and 13 ToxiGen target groups, showing that adversarial vulnerability and suppression effectiveness vary substantially across groups, exposing structural fairness gaps in current toxicity classifiers.
\end{itemize}

These findings inform a principled, proactive approach to hardening toxicity classifiers that is more trustworthy, transparent, and equitable.

\section{\textbf{Related Work}}

\subsection{\textbf{Toxicity Detection}}
Toxicity detection has been widely studied across languages \cite{khan2024abusive,bensalem2024toxic,krak2024abusive,jiang2024cross,mnassri2024survey}. Early keyword and bag-of-words systems gave way to transformer-based models like BERT, which brought substantial gains on benchmarks like Jigsaw \cite{jigsaw-unintended-bias-in-toxicity-classification,nobata2016abusive,wulczyn2017ex,yin2021towards}. Recent datasets targeting implicit or subtle toxicity include ETHOS \cite{mollas_ethos_2022} and ToxiGen \cite{toxigen}, and multi-label setups now distinguish threats, insults, and identity attacks \cite{casellietal2020}. Despite these advances, classifiers remain brittle under distributional shift, over-relying on lexical cues and exhibiting demographic disparities -- marginalised-identity comments are disproportionately flagged as toxic \cite{yin2021towards,dixon2018measuring}. Proposed remedies include counterfactual augmentation, adversarial training, and fairness-aware losses \cite{dinan2020multi,samory_sexism}, but most evaluations use static splits that do not capture adversarially manipulated text, motivating our mechanistic analysis.



\subsection{\textbf{LLM-Generated Content and Adversarial Attacks}}
With the advent of LLMs and the associated ease to produce LLM-generated content at scale, a new need has emerged for building classification models that identify if a text is human- or LLM-generated \cite{wu-etal-2025-survey}. Research in LLM-generated text detection has evolved from development of classifiers that rely on linguistic features \cite{frohling2021feature}, to developing watermarking techniques to facilitate detection \cite{pan2024markllm}, statistics-based detectors \cite{vasilatos2023howkgpt}, and neural-based detectors \cite{mitchell2023detectgpt}. Despite increasing interest in detecting LLM-generated text and identifying what distinguishes it from human text, the use of LLMs for generation of toxic content has been comparatively understudied.

The rapid adoption of LLMs has led to large volumes of synthetic toxic text; because classifiers are trained on human-written data, LLM outputs often fall outside their distribution \cite{BARMAN2024100545, Warner_2025}. Adversarial prompting further enables semantically-preserved toxic content that circumvents detection \cite{guo2021gradientbasedadversarialattackstext, tokenmodification, Sadrizadeh2023ACA, Li_2019}. Prior defenses -- adversarial training \cite{bespalov2024buildingrobusttoxicitypredictor, samory_sexism} or detection filters \cite{chen2022should, yoo-etal-2022-detection} -- remain reactive, requiring continuous retraining as attacks evolve.

\begin{figure*}[htb]
\centering
\resizebox{\textwidth}{!}{%
\begin{tikzpicture}[
  font=\sffamily,
  >={Latex[length=2.2mm,width=2.0mm]},
  line join=round,
  stage/.style={rectangle, rounded corners=3pt, thick,
                draw=#1!65!black, fill=#1!6,
                minimum width=52mm, minimum height=75mm,
                inner sep=0pt},
  setupS/.style   ={stage=blue},
  attackS/.style  ={stage=red},
  patchS/.style   ={stage=orange},
  suppressS/.style={stage=green!55!black},
  hdr/.style={rectangle, rounded corners=2pt, thick,
              draw=#1!65!black, fill=#1!22,
              inner sep=2.5pt, font=\bfseries\sffamily\small},
  card/.style={rectangle, rounded corners=2pt, draw=black!35, fill=white,
               inner sep=3pt, font=\scriptsize\sffamily, align=center,
               text width=45mm},
  eqcard/.style={card, font=\scriptsize\sffamily, align=center},
  gridcard/.style={rectangle, rounded corners=2pt, draw=black!35, fill=white,
                   inner sep=3pt, align=center, text width=45mm,
                   minimum height=18mm},
  mainarr/.style={->, line width=0.9pt, black!75},
  arr/.style={->, line width=0.6pt, black!70},
  arrlab/.style={font=\scriptsize\sffamily\itshape, text=black!65,
                 fill=white, inner sep=1.5pt, align=center},
  hcell/.style={rectangle, minimum size=1.15mm, inner sep=0pt, draw=black!28,
                fill=white},
  hhot/.style ={hcell, fill=red!75!black, draw=red!80!black},
  hwarm/.style={hcell, fill=orange!65!yellow, draw=orange!75!black},
  hcool/.style={hcell, fill=blue!55,  draw=blue!70}
]

\node[setupS] (S1) at (0,0) {};
\node[hdr=blue, anchor=north, yshift=-2mm] (H1) at (S1.north)
     {1.~Setup \& Fine-tuning};

\node[card, anchor=north, yshift=-10mm] (corp) at (S1.north) {%
  \textbf{Corpora}\\[1pt]
  Jigsaw: 50k human (Wikipedia)\\
  ToxiGen: 50k LLM-generated\\
  10k balanced eval, per corpus};
\node[card, below=2mm of corp] (bert) {%
  \textbf{BERT-base}\\
  \texttt{bert-base-uncased}\\
  110M, 12L$\times$12H = 144 heads};
\node[card, below=2mm of bert] (rob) {%
  \textbf{RoBERTa-base}\\
  \texttt{roberta-base}\\
  125M, 12L$\times$12H = 144 heads};
\node[card, below=2mm of rob] (lg) {%
  \textbf{Llama Guard 2}\\
  \texttt{Meta-Llama-Guard-2-8B}\\
  8B, 32L$\times$32H = 1024 heads};

\node[attackS, right=12mm of S1] (S2) {};
\node[hdr=red, anchor=north, yshift=-2mm] (H2) at (S2.north)
     {2.~PGD BERT-Attack};

\node[card, anchor=north, yshift=-10mm] (xclean) at (S2.north) {%
  \textbf{clean input} $x$\\
  \texttt{"i hate [group], they\ldots"}};
\node[eqcard, below=2mm of xclean] (emb) {%
  \textbf{Embed:}~$E_0=E(x)$};
\node[eqcard, below=2mm of emb] (pgd) {%
  \textbf{PGD loop}~($t=1{:}T$)\\[1pt]
  $g_t=\nabla_{\!E}\,\mathcal{L}\bigl(f_\theta(E_{t-1}),y\bigr)$\\
  $E_t=\Pi_{\|\cdot\|_\infty\le\varepsilon}\!\bigl(E_{t-1}+\alpha\,g_t\bigr)$};
\node[eqcard, below=2mm of pgd] (snap) {%
  \textbf{Snap to tokens:}\\
  $x_{\text{adv}}=\arg\min_{w}\|E_T-E(w)\|$};
\node[card, below=2mm of snap] (xadv) {%
  \textbf{adversarial} $x_{\text{adv}}$\\
  \texttt{"i h8 [grp], th3y\ldots"}};

\draw[arr] (xclean) -- (emb);
\draw[arr] (emb)    -- (pgd);
\draw[arr] (pgd)    -- (snap);
\draw[arr] (snap)   -- (xadv);

\node[patchS, right=12mm of S2] (S3) {};
\node[hdr=orange, anchor=north, yshift=-2mm] (H3) at (S3.north)
     {3.~Activation Patching};

\node[gridcard, anchor=north, yshift=-10mm] (clp) at (S3.north) {};
\node[anchor=north, yshift=-1mm,
      font=\scriptsize\sffamily\bfseries, text=red!60!black]
     at (clp.north) {Clean input $x$};
\node[anchor=south, yshift=1mm, font=\scriptsize\sffamily]
     at (clp.south) {$\Delta\mathcal{L}_{\text{clean}}\!\uparrow\Rightarrow$ \emph{crucial}};
\node[anchor=center] (aC) at ([yshift=-0.5mm]clp.center) {};
\begin{scope}[shift={(aC)}]
  \foreach \r in {0,...,11}
    \foreach \c in {0,...,11}
      \node[hcell] at (\c*1.15mm-6.3mm, -\r*1.15mm+6.3mm) {};
  \node[hhot]  at (5*1.15mm-6.3mm,-9*1.15mm+6.3mm) {}; 
  \node[hhot]  at (1*1.15mm-6.3mm,-6*1.15mm+6.3mm) {}; 
  \node[hwarm] at (8*1.15mm-6.3mm,-10*1.15mm+6.3mm) {};
\end{scope}

\node[gridcard, below=2mm of clp] (advp) {};
\node[anchor=north, yshift=-1mm,
      font=\scriptsize\sffamily\bfseries, text=blue!60!black]
     at (advp.north) {Adversarial $x_{\text{adv}}$};
\node[anchor=south, yshift=1mm, font=\scriptsize\sffamily]
     at (advp.south) {$\Delta\mathcal{L}_{\text{adv}}\!\downarrow\Rightarrow$ \emph{vulnerable}};
\node[anchor=center] (aA) at ([yshift=-0.5mm]advp.center) {};
\begin{scope}[shift={(aA)}]
  \foreach \r in {0,...,11}
    \foreach \c in {0,...,11}
      \node[hcell] at (\c*1.15mm-6.3mm, -\r*1.15mm+6.3mm) {};
  \node[hcool] at (5*1.15mm-6.3mm,-9*1.15mm+6.3mm) {}; 
  \node[hcool] at (1*1.15mm-6.3mm,-6*1.15mm+6.3mm) {}; 
  \node[hwarm] at (4*1.15mm-6.3mm,-8*1.15mm+6.3mm) {};
  \node[hwarm] at (7*1.15mm-6.3mm,-7*1.15mm+6.3mm) {};
\end{scope}

\node[card, below=2mm of advp, text width=45mm] (inter) {%
  $\mathcal{H}^{\star}{=}\mathrm{top}\text{-}2\bigl(\text{crucial}\cap\text{vuln.}\bigr)$\\[1pt]
  {\scriptsize e.g.\ $\mathcal{H}^{\star}_{\mathrm{Jigsaw}}{=}\{L9H5,\,L6H1\}$}};

\draw[arr] (clp)  -- (advp);
\draw[arr] (advp) -- (inter);

\node[suppressS, right=12mm of S3] (S4) {};
\node[hdr=green!55!black, anchor=north, yshift=-2mm] (H4) at (S4.north)
     {4.~Suppress \& Re-evaluate};

\node[card, anchor=north, yshift=-10mm, text width=45mm] (abl) at (S4.north) {%
  \textbf{Ablate $\mathcal{H}^{\star}$}\\
  zero head outputs:\\
  $\tilde f_\theta=f_\theta \setminus \mathcal{H}^{\star}$};

\node[card, below=2mm of abl] (m1) {%
  \textbf{Adversarial acc}~\textcolor{green!45!black}{$\uparrow$}\\[1pt]
  re-evaluate $\tilde f_\theta$ on $x_{\text{adv}}$};
\node[card, below=2mm of m1]  (m2) {%
  \textbf{Clean acc}~$\approx$\\[1pt]
  natural 7.9\%-toxic eval};
\node[card, below=2mm of m2]  (m3) {%
  \textbf{Demographic gap}~\textcolor{green!45!black}{$\downarrow$}\\[1pt]
  per-group adv.\ acc.\ narrows};
\node[card, below=2mm of m3]  (m4) {%
  \textbf{Validation}\\[1pt]
  split-sample + class-imbalance sweep};

\draw[mainarr] (abl) --
    node[arrlab, right=1pt, fill=none, text=black!60] {outcomes}
    (m1);

\draw[mainarr] (S1.east |- H1) --
   node[arrlab, above] {$f_\theta,\,x,\,y$}
   (S2.west |- H2);
\draw[mainarr] (S2.east |- H2) --
   node[arrlab, above] {$(x,\,x_{\text{adv}})$}
   (S3.west |- H3);
\draw[mainarr] (S3.east |- H3) --
   node[arrlab, above] {$\mathcal{H}^{\star}$}
   (S4.west |- H4);

\draw[arr, dashed, black!55]
  (S4.south) .. controls +(0,-10mm) and +(0,-10mm) ..
  node[arrlab, below=1pt, fill=none, text=black!55]
     {evaluate $\tilde f_\theta$ on $(x,\,x_{\text{adv}})$}
  (S2.south);

\end{tikzpicture}}
\caption{End-to-end pipeline from adversarial attack to targeted head suppression across six model--corpus combinations. Fine-tuned classifiers (\textbf{Stage~1}) are attacked via PGD BERT-Attack (\textbf{Stage~2}); activation patching identifies the bottleneck head set $\mathcal{H}^{\star}$ (\textbf{Stage~3}); zero-ablating $\mathcal{H}^{\star}$ improves adversarial accuracy while preserving clean performance (\textbf{Stage~4}). The dashed arc denotes evaluation of the suppressed model $\tilde f_\theta$ on both clean and adversarial inputs.}
\label{fig:pipeline_overview}
\end{figure*}

\subsection{\textbf{Mechanistic Interpretability for Robustness}}
Mechanistic interpretability (MI) aims to identify functional circuits within neural networks, enabling a causal understanding of model behavior \cite{Olah2020,bereska2024mechanistic}. Activation patching and causal tracing techniques have revealed specialized circuits in tasks such as syntactic role identification \cite{wang2022interpretabilitywildcircuitindirect}, arithmetic reasoning \cite{hanna2023does}, and factual recall \cite{conmy2023automatedcircuitdiscoverymechanistic}. Recently, \citet{Garc_a_Carrasco_2024} extended these methods to systematically discover circuits in smaller classification settings, demonstrating their potential for principled interventions. However, applications of MI to toxicity classification or adversarial robustness remain unexplored. By analyzing which attention heads are crucial for performance and which are vulnerable to adversarial perturbations, our work connects mechanistic interpretability with robustness research, providing a proactive strategy to mitigate attacks.

In contrast to prior work, we combine three threads: (1) the challenge of classifying both human- and LLM-generated toxic text, (2) the need for defenses that generalize across unseen adversarial attacks, and (3) the promise of mechanistic interpretability to expose internal model failure modes. Our approach identifies vulnerable circuits in transformer-based classifiers and suppresses them, improving robustness while revealing fairness gaps across demographic groups.

\section{\textbf{Methodology}}

To identify the components most vulnerable to misclassification, we follow the mechanistic interpretability (MI) workflow \cite{conmy2023automatedcircuitdiscoverymechanistic, Garc_a_Carrasco_2024}, previously applied to indirect object identification \cite{wang2022interpretabilitywildcircuitindirect} and mathematical reasoning \cite{hanna2023does}. Figure~\ref{fig:pipeline_overview} summarizes our eight-phase pipeline from attack generation through head localization to targeted suppression and re-evaluation.

We apply the full pipeline to a \textbf{2$\times$2 factorial design}: BERT \cite{devlin2019bertpretrainingdeepbidirectional} (\texttt{bert-base-uncased}, 110M params, 12 layers $\times$ 12 heads = 144 heads) and RoBERTa \cite{liu2019roberta} (\texttt{roberta-base}, 125M params, 12 layers $\times$ 12 heads = 144 heads), each fine-tuned independently on Jigsaw and ToxiGen. We additionally extend the core patching experiments to \textbf{Llama Guard~2} \cite{inan2023llamaguard} (\texttt{meta-llama/Meta-Llama-Guard-2-8B}), a 32-layer, 32-head decoder-only LLM (1,024 heads total) that outputs \texttt{safe}/\texttt{unsafe} as its first generated token. All fine-tuned models are trained from HuggingFace weights with no pre-existing task-specific checkpoint.

\subsection{\textbf{Datasets and Metrics}}

We use the following two datasets:

\paragraph{ToxiGen} \cite{toxigen}. A machine-generated English dataset of 250k statements covering 13 minority groups, generated by GPT-3. We use 50,000 stratified training samples (90\% split) and 10,000 balanced test samples (5,000/class). For PGD we use 7,400 samples ($\approx$50\% toxic). Demographic analysis uses the \texttt{annotated} config: 940 examples across 13 groups (Asian, Black/African-American, Chinese, Mental/Physical Disability, Jewish, Latino/Hispanic, LGBTQ+, Mexican, Middle Eastern, Muslim, Native American/Indigenous, Women).

\paragraph{Jigsaw} \cite{jigsaw-unintended-bias-in-toxicity-classification}. Human-written Wikipedia comments. We use 50,000 stratified training samples and two test sets: 10,000 balanced examples (5,000/class) for evaluation, and 7,400 natural-distribution samples for PGD (583 toxic / 6,817 non-toxic, 7.9\% toxic). Demographic analysis uses the full 50,000-sample pool annotated with 20 groups (Asian, Atheist, Bisexual, Black, Buddhist, Christian, Female, Heterosexual, Hindu, Gay or Lesbian, Intellectual/Learning Disability, Jewish, Latino, Male, Muslim, Other, Physical Disability, Psychiatric/Mental Illness, Transgender, White).

We evaluate using binary cross-entropy loss $\mathcal{L} = -[ y \log p + (1{-}y) \log(1{-}p) ]$ and accuracy (fraction of labels correctly predicted).

\subsection{\textbf{Step 1. Identifying Crucial Heads Using Activation Patching}}

Activation patching replaces each head's output with zeros \cite{olsson2022context} and measures the change in loss. If zeroing a head substantially decreases performance, that head is \textit{crucial} to the task. We perform this sequentially across all 144 heads (or 1,024 for Llama Guard~2) on the clean test set.

\begin{figure*}[htb]
    \centering
    \includegraphics[width=\linewidth]{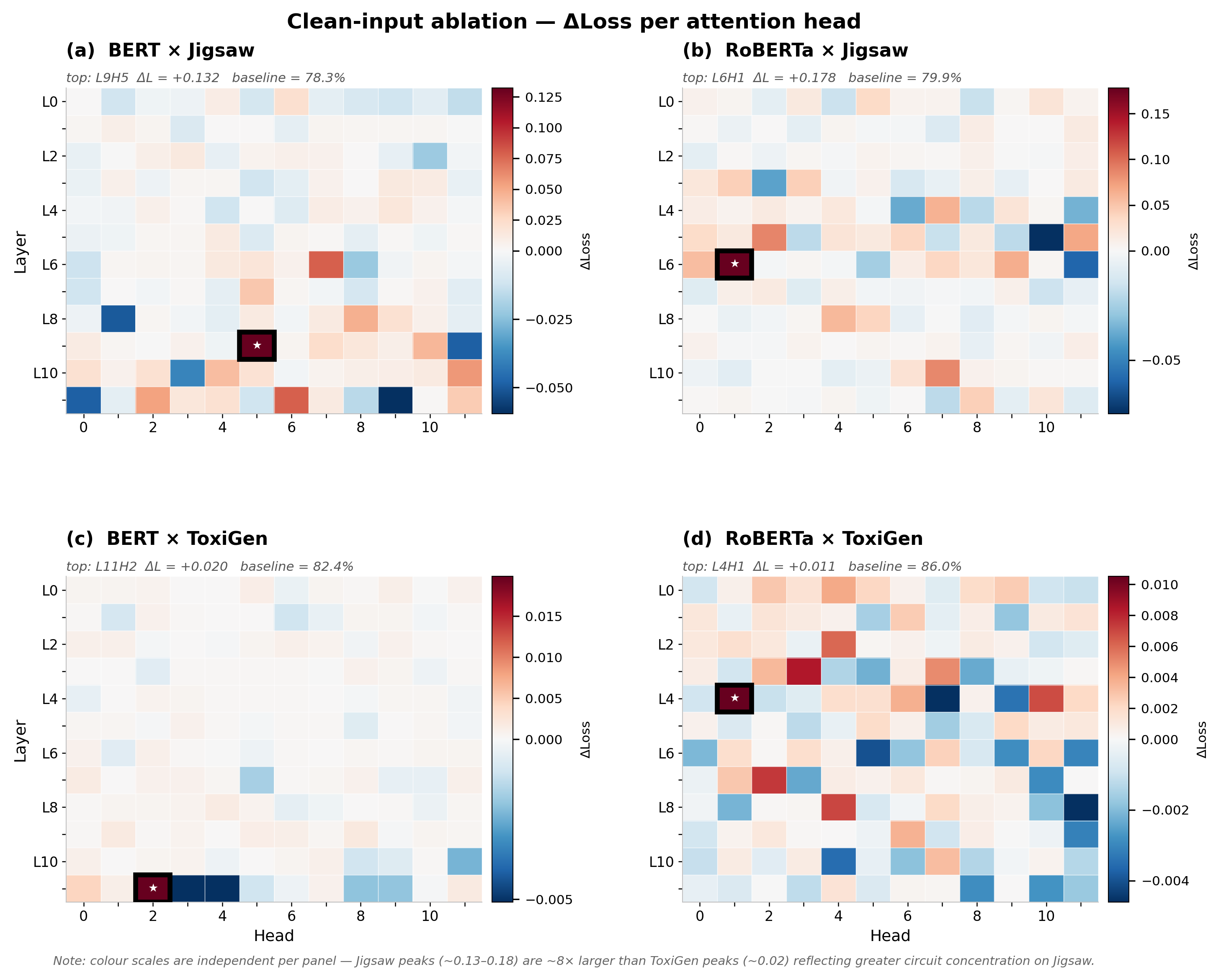}
    \caption{Clean-input activation patching: per-head $\Delta$Loss (increase = head is crucial) across all four runs in a 2$\times$2 grid (rows: BERT/RoBERTa; columns: Jigsaw/ToxiGen). Warmer colours indicate higher impact. Jigsaw models show a single dominant hot cell (L9H5, L6H1); ToxiGen models are diffuse.}
    \label{fig:clean_heatmaps}
\end{figure*}

\subsection{\textbf{Step 2. Adversarial Attacks}}

\paragraph{PGD BERT-Attack} \cite{waghela2024enhancingadversarialtextattacks}. A white-box attack that iteratively perturbs input embeddings via gradient ascent within a bounded $\ell_\infty$ norm ball, then snaps to the nearest valid tokens. This yields worst-case token substitutions that are semantically imperceptible (Table~\ref{tab:adv_example_transposed}); the full pipeline is in Figure~\ref{fig:bert_attack_diagram} (Appendix~\ref{app:figs}).

\subsection{\textbf{Step 3. Identifying Vulnerable Heads on Adversarial Input}}

Using the adversarially-attacked dataset, we repeat activation patching across all heads. Heads whose ablation \textit{reduces} loss on adversarial inputs are \textit{vulnerable} -- complicit in the attack's success.

\subsection{\textbf{Step 4. Targeted Head Suppression and Re-evaluation}}

Given the ranked vulnerable head set from Step~3, we construct the suppressed model $\tilde{f}_\theta = f_\theta \setminus \mathcal{H}^{\star}$ by zeroing the outputs of the top-$k$ vulnerable heads at inference time -- no retraining required. We then re-evaluate $\tilde{f}_\theta$ on both clean and adversarial inputs. The key quantities are: (i)~adversarial accuracy recovery ($\Delta$ adv acc), measuring how much of the near-zero baseline is restored; (ii)~clean accuracy cost ($\Delta$ clean acc), measuring the toll on non-adversarial performance; and (iii)~demographic-level recovery across all minority groups. We additionally compare suppression against two retraining baselines -- FGM adversarial training and data augmentation -- to assess whether an interpretability-derived, zero-cost intervention can match approaches that require full retraining.

\section{\textbf{Experimental Setup}}

All experiments use dedicated H100 GPUs (one per run).

\subsection{\textbf{Model Training}}

All four models (bert\_jigsaw, roberta\_jigsaw, bert\_toxigen, roberta\_toxigen) are fine-tuned from HuggingFace weights for \textbf{3 epochs} with batch size 512, AdamW optimiser, and a cosine learning-rate schedule. Evaluation uses the \textbf{10,000-sample balanced test set} (5,000 toxic / 5,000 non-toxic). Table~\ref{tab:training} summarises the final checkpoint accuracies.

\begin{table}[h]
\centering
\small
\begin{tabular}{llr}
\toprule
\textbf{Run} & \textbf{Architecture} & \textbf{Clean Acc (balanced)} \\
\midrule
bert\_jigsaw     & BERT     & 78.35\% \\
roberta\_jigsaw  & RoBERTa  & 79.88\% \\
bert\_toxigen    & BERT     & 82.40\% \\
roberta\_toxigen & RoBERTa  & 85.98\% \\
\bottomrule
\end{tabular}
\caption{Best checkpoint accuracy on the 10k balanced test set after 3 training epochs.}
\label{tab:training}
\end{table}

On Jigsaw's natural 7.9\%-toxic distribution all models reach 93--95\% accuracy; the balanced 10k set is used for all mechanistic comparisons below.

\subsection{\textbf{Adversarial Attack Procedure}}

We generate adversarial examples using PGD BERT-Attack \cite{waghela2024enhancingadversarialtextattacks} over a 7,400-sample test set per run (24 workers, 10 iterations, \texttt{bert-base-uncased} as MLM). Clean-set success rates are: bert\_jigsaw 67.9\% (5,026 examples), roberta\_jigsaw 32.6\% (2,415), bert\_toxigen 69.1\% (5,116), roberta\_toxigen 60.8\% (4,496). Demographic-pool attacks yield 33,894 / 16,438 / 717 / 674 adversarial examples respectively. RoBERTa Jigsaw's low success rate (32.6\%, less than half BERT's) reflects a substantial architecture effect on attack vulnerability.

\section{\textbf{Results}}

Figure~\ref{fig:clean_heatmaps} shows per-head $\Delta$Loss on clean input across all four runs; adversarial and demographic heatmaps are in Appendix~\ref{app:figs}. Figure~\ref{fig:side_by_side_acc_change} shows per-group adversarial accuracy before and after top-2 suppression.

\subsection{\textbf{Phase 1: Fine-Tuning and Adversarial Baseline}}

\paragraph{\textbf{Adversarial attacks reduce all models to near-zero accuracy.}}
PGD attacks bring every model to near-complete failure on adversarial inputs: 0.06\% (bert\_jigsaw), 0.21\% (roberta\_jigsaw), 0.20\% (bert\_toxigen), and 0.09\% (roberta\_toxigen) adversarial accuracy, from clean baselines of 78--86\%. This confirms that even the strongest fine-tuned classifiers are entirely vulnerable to white-box token-substitution attacks.

\subsection{\textbf{Phase 2: Clean Activation Patching}}

We zero-ablate each of 144 attention heads on 10,000 clean test examples and rank by increase in cross-entropy loss ($\Delta$Loss). Table~\ref{tab:clean_heads} shows the top head per run.

\begin{table}[tb]
\centering
\small
\begin{tabular}{llrr}
\toprule
\textbf{Run} & \textbf{Top head} & \textbf{$\Delta$Loss} & \textbf{Acc drop} \\
\midrule
bert\_jigsaw     & L9H5  & +0.132 & $-$5.03 pp \\
roberta\_jigsaw  & \textbf{L6H1}  & +0.178 & $-$\textbf{11.58 pp} \\
bert\_toxigen    & L11H2 & +0.020 & $-$0.55 pp \\
roberta\_toxigen & L4H1  & +0.011 & $-$0.54 pp \\
\bottomrule
\end{tabular}
\caption{Top crucial head per run identified by clean activation patching.}
\label{tab:clean_heads}
\end{table}

\paragraph{\textbf{Jigsaw classifiers rely on a single dominant head; ToxiGen classifiers are distributed.}}
Jigsaw models concentrate clean classification into one head: ablating L9H5 alone costs bert\_jigsaw 5 pp, while ablating L6H1 alone costs roberta\_jigsaw \textbf{11.58 pp}, dropping it from 79.88\% to 68.30\%. The $\Delta$Loss magnitudes for Jigsaw's top heads (0.132, 0.178) are \textbf{7--15$\times$ larger} than those for ToxiGen (0.020, 0.011). ToxiGen crucial heads are distributed across many layers with no single bottleneck, consistent with ToxiGen requiring more nuanced processing of subtle, generated hate speech.

\subsection{\textbf{Phase 3: Adversarial Activation Patching}}

We zero-ablate each head on the successfully-attacked adversarial examples and rank by \textit{decrease} in cross-entropy loss (negative $\Delta$Loss; the head is complicit in adversarial vulnerability). Table~\ref{tab:adv_heads} reports single-head and cumulative 5-head adversarial accuracy recovery.

\begin{table}[tb]
\centering
\small
\begin{tabular}{llrrr}
\toprule
\textbf{Run} & \textbf{Top head} & \textbf{$\Delta$Loss} & \textbf{k=1} & \textbf{k=5} \\
\midrule
bert\_jigsaw     & L9H5 & $-$0.562 & +50.5 pp & +84.9 pp \\
roberta\_jigsaw  & \textbf{L6H1} & $-$\textbf{0.721} & \textbf{+70.4 pp} & +76.7 pp \\
bert\_toxigen    & L9H5 & $-$0.069 & +23.6 pp & +34.1 pp \\
roberta\_toxigen & L6H1 & $-$0.174 & +23.5 pp & +38.9 pp \\
\bottomrule
\end{tabular}
\caption{Top vulnerable head and adversarial accuracy recovery ($\Delta$ from near-zero baseline) at $k$=1 and $k$=5 heads suppressed.}
\label{tab:adv_heads}
\end{table}

\paragraph{\textbf{The same head is dominant for both clean and adversarial circuits on Jigsaw.}}
L9H5 is rank~1 in \textit{both} clean and adversarial patching for bert\_jigsaw; L6H1 is rank~1 in both for roberta\_jigsaw. This means Jigsaw classification is mediated through a single bottleneck circuit that PGD learns to exploit. In contrast, for ToxiGen models the clean-critical and adversarially-vulnerable heads are largely \textit{disjoint}: a separate early-layer circuit is hijacked by the attack.

\begin{figure*}[htb]
    \centering
    \includegraphics[width=\linewidth]{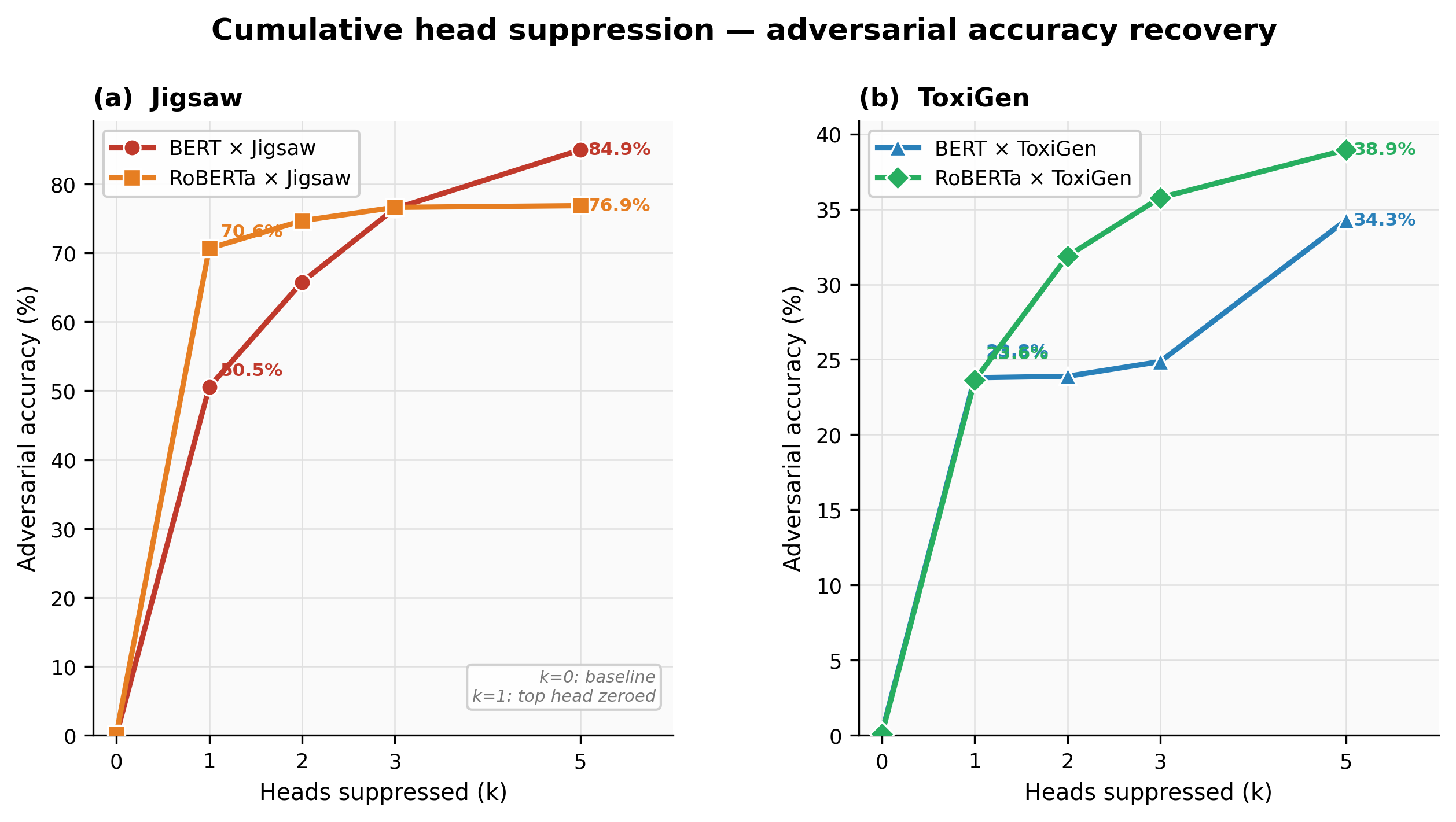}
    \caption{Adversarial accuracy recovery as $k$ heads are suppressed. Jigsaw curves (left) rise steeply at $k$=1; ToxiGen curves (right) are shallower across all $k$.}
    \label{fig:suppression_curves}
\end{figure*}

Three profiles emerge (Figure~\ref{fig:suppression_curves}): extreme single-head dominance for RoBERTa Jigsaw (70.4 pp at $k$=1, +6.2 pp more at $k$=5); strong bottleneck with additivity for BERT Jigsaw (50.5$\rightarrow$84.9 pp); and moderate distributed vulnerability for ToxiGen ($\approx$23--24 pp, requiring more heads for meaningful recovery).

\paragraph{\textbf{The same architecture-level bottleneck recurs across datasets.}}
A striking convergence is visible in Table~\ref{tab:adv_heads}: the rank-1 adversarially vulnerable head is \emph{identical across both datasets within each architecture} -- L9H5 for all BERT runs, L6H1 for all RoBERTa runs. Despite ToxiGen's distributed processing signature, PGD exploits the same structural bottlenecks, albeit with 8--10$\times$ smaller $\Delta$Loss magnitudes (BERT: $-$0.562 Jigsaw vs $-$0.069 ToxiGen; RoBERTa: $-$0.721 vs $-$0.174). The circuit bottleneck is an architecture-level property; the model concentrates more functional load onto it for the more lexically predictable Jigsaw task, amplifying exploitability.

\subsection{\textbf{Phase 4: Demographic-Stratified Adversarial Analysis}}

We suppress the top-2 vulnerable heads per run on adversarial examples from the demographic pool (bert\_jigsaw: 33,894; roberta\_jigsaw: 16,438; bert\_toxigen: 717; roberta\_toxigen: 674 examples). Table~\ref{tab:demo_overall} summarises overall effectiveness; per-group bar charts are in Figure~\ref{fig:side_by_side_acc_change}, and per-head group heatmaps in Appendix~\ref{app:figs}.

\begin{table}[tb]
\centering\small
\setlength{\tabcolsep}{3pt}
\resizebox{\columnwidth}{!}{%
\begin{tabular}{llrrr}
\toprule
\textbf{Run} & \textbf{Heads} & \textbf{Base} & \textbf{Supp.} & \textbf{$\Delta$} \\
\midrule
bert\_jigsaw     & L9H5, L8H8   & 0.17\% & 67.22\% & +67.1 pp \\
roberta\_jigsaw  & L6H1, L5H2   & 0.09\% & 56.79\% & +56.7 pp \\
bert\_toxigen    & L9H5, L11H3  & 0.16\% & 25.66\% & +25.5 pp \\
roberta\_toxigen & L6H1, L4H10  & 0.33\% & 27.46\% & +27.1 pp \\
\bottomrule
\end{tabular}}
\caption{Overall adversarial accuracy before and after top-2 head suppression on the demographic adversarial pool.}
\label{tab:demo_overall}
\end{table}

\begin{figure*}[htb]
    \centering
    \begin{subfigure}{0.48\textwidth}
        \centering
        \includegraphics[width=\linewidth]{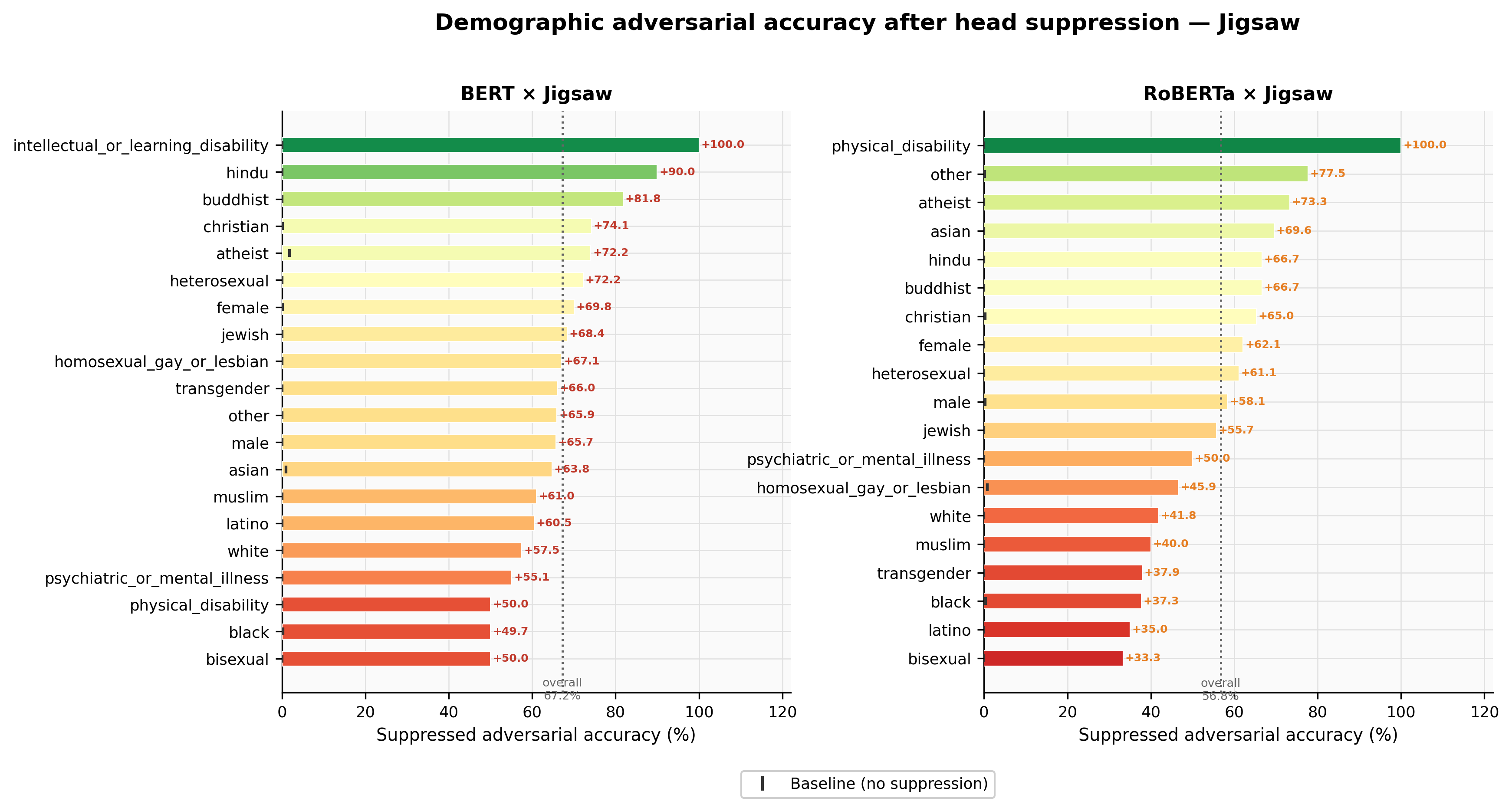}
        \caption{Jigsaw (20 groups).}
        \label{fig:jigsaw_dem_acc_change}
    \end{subfigure}
    \hfill
    \begin{subfigure}{0.48\textwidth}
        \centering
        \includegraphics[width=\linewidth]{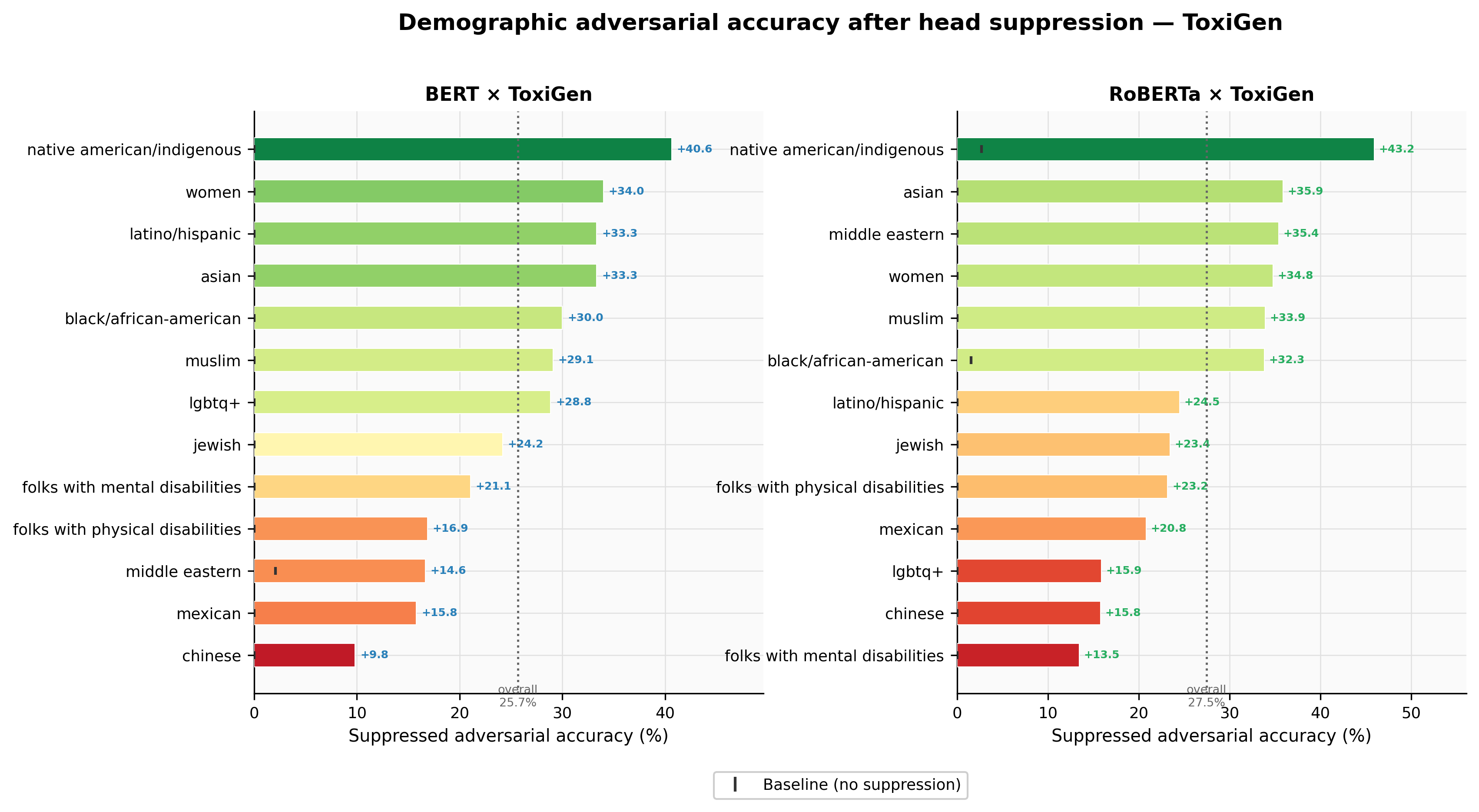}
        \caption{ToxiGen (13 groups).}
        \label{fig:toxigen_dem_acc_change}
    \end{subfigure}
    \caption{Per-group adversarial accuracy after top-2 head suppression. Groups sorted by suppressed accuracy; $*$ = $n < 30$.}
    \label{fig:side_by_side_acc_change}
\end{figure*}

\begin{figure*}[htb]
    \centering
    \includegraphics[width=\linewidth]{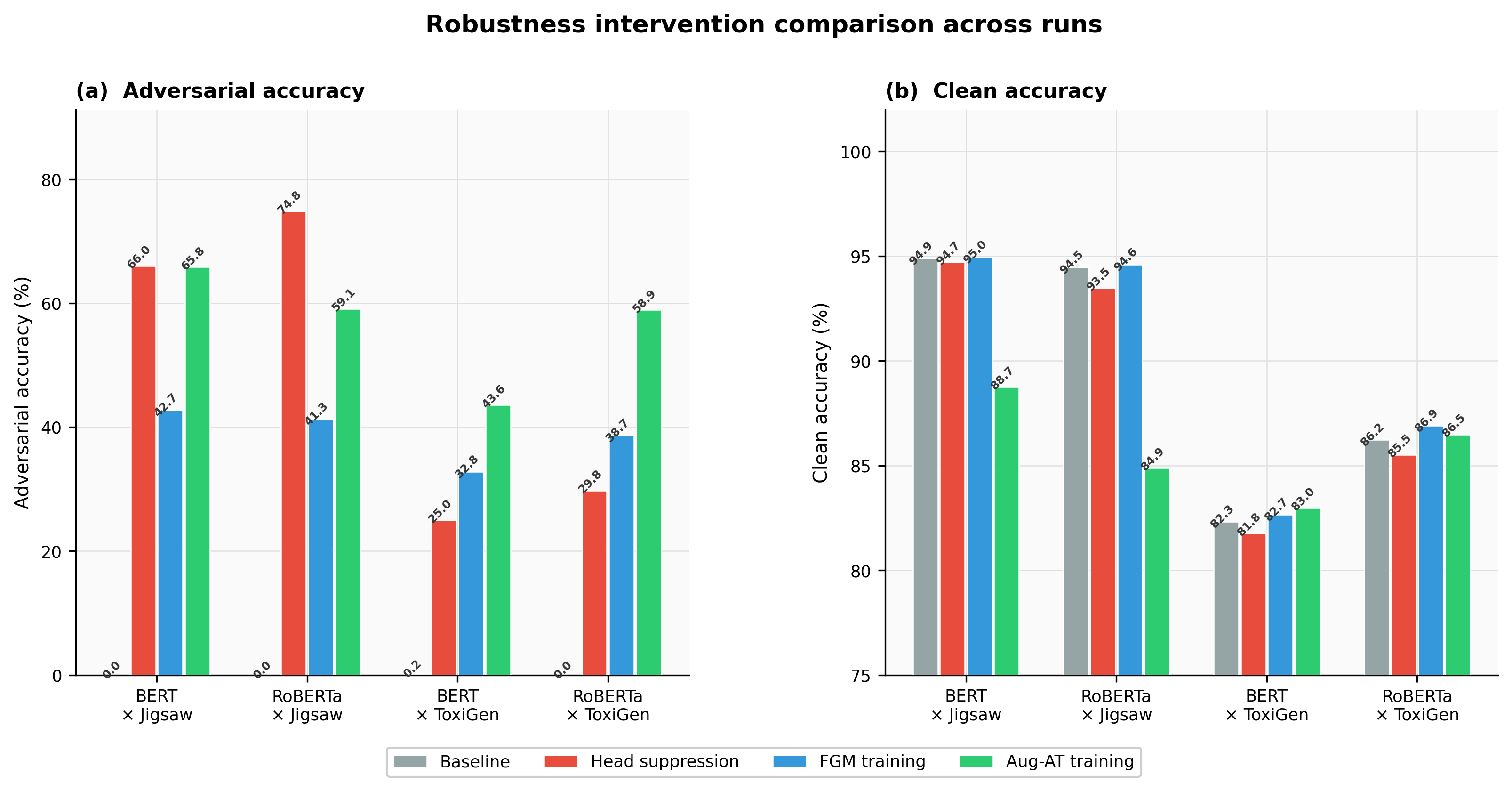}
    \caption{Adversarial (solid) and clean (hatched) accuracy for all robustness methods. Head suppression wins on Jigsaw; augmentation wins on ToxiGen.}
    \label{fig:robustness_bars}
\end{figure*}

\paragraph{\textbf{Adversarial vulnerability is distributed unequally across demographic groups.}}
Jigsaw head suppression recovers 2.5$\times$ more adversarial accuracy than ToxiGen, consistent with stronger single-head bottlenecks. Within each dataset, recovery varies dramatically: for BERT Jigsaw, Buddhist (+81.8 pp), Hindu (+90.0 pp), and Intellectual/Learning Disability (\textbf{+100 pp}) are highest; Black (+49.7 pp), Bisexual (+50.0 pp), and White (+57.5 pp) are lower. For ToxiGen, Native American/Indigenous is the consistently highest-delta group (+40.6 pp BERT, +43.2 pp RoBERTa), while Chinese and disability-related groups show the weakest recovery.

Head-level group heatmaps (Appendix~\ref{app:figs}) show that certain heads are selectively exploited for specific communities, evidence that demographic-specific adversarial bias is mechanistically encoded into individual heads.

\paragraph{\textbf{Disability groups show a cross-dataset circuit split.}}
Disability-related groups reveal a striking cross-dataset divergence. In ToxiGen, \textit{mental disability} and \textit{physical disability} groups rank at the bottom of suppression response across both models, suggesting that the adversarial circuit for disability-targeted LLM-generated content is not concentrated in L9H5/L6H1. In Jigsaw, the pattern inverts: \textit{intellectual\_or\_learning\_disability} achieves \textbf{100\%} suppressed accuracy (BERT) and \textit{psychiatric\_or\_mental\_illness} achieves 50--55\%, both above the overall mean. This divergence suggests the adversarial representation of disability content engages distinct head circuits depending on whether the text is human-written (Jigsaw) or LLM-generated (ToxiGen), with implications for dataset-targeted fairness interventions. Note: several Jigsaw groups (Buddhist, Hindu, Bisexual, ILD, Physical Disability) have $n{<}30$ adversarial examples; extreme values (90--100\% recovery) should be interpreted cautiously.

\subsection{\textbf{Phase 5: Adversarial Robustness Interventions}}

We compare four conditions: (1) baseline, (2) top-2 head suppression at inference, (3) FGM adversarial training ($\varepsilon=1.0$, 10 epochs), (4) data augmentation training (clean:adv ratio 3:1, 10 epochs). Clean accuracy is evaluated on the natural (unbalanced) test distribution.

\begin{table}[h]
\centering
\small
\setlength{\tabcolsep}{4pt}
\begin{tabular}{llrr}
\toprule
\textbf{Run} & \textbf{Method} & \textbf{Clean} & \textbf{Adv} \\
\midrule
bert\_jig.   & Baseline       & 94.88\% & 0.00\% \\
             & Head supp.     & 94.71\% & \textbf{66.00\%} \\
             & FGM            & 94.96\% & 42.74\% \\
             & Augmentation   & 88.74\% & 65.81\% \\
\midrule
rob.\_jig.   & Baseline       & 94.45\% & 0.00\% \\
             & Head supp.     & 93.89\% & \textbf{73.14\%} \\
             & FGM            & 94.89\% & 69.01\% \\
             & Augmentation   & 83.56\% & 59.09\% \\
\midrule
bert\_tox.   & Baseline       & 82.33\% & 0.20\% \\
             & Head supp.     & 81.36\% & 29.10\% \\
             & FGM            & 83.16\% & 34.96\% \\
             & Augmentation   & 82.87\% & \textbf{45.31\%} \\
\midrule
rob.\_tox.   & Baseline       & 86.24\% & 0.00\% \\
             & Head supp.     & 85.52\% & 29.78\% \\
             & FGM            & 86.90\% & 38.67\% \\
             & Augmentation   & 86.48\% & \textbf{58.89\%} \\
\bottomrule
\end{tabular}
\caption{Clean and adversarial accuracy for all robustness methods. \textbf{Bold} = best adv.\ accuracy per run. Clean accuracy on unbalanced natural test.}
\label{tab:interventions}
\end{table}

\paragraph{\textbf{The best robustness method reverses between datasets.}}
For Jigsaw, \textbf{head suppression matches or outperforms all adversarial training approaches} at a cost of only $-$0.2 to $-$0.6 pp clean accuracy. Augmentation training achieves comparable adversarial accuracy on Jigsaw but at a severe clean penalty ($-$6.1 pp BERT, $-$10.9 pp RoBERTa). For ToxiGen, \textbf{augmentation training dominates} (+45.3 pp BERT, +58.9 pp RoBERTa adversarial accuracy), while head suppression is weakest. Notably, ToxiGen augmentation incurs a \emph{negligible} clean penalty (+0.5 pp BERT, +0.2 pp RoBERTa), in sharp contrast to Jigsaw's severe cost. This asymmetry directly supports the bottleneck/distributed dichotomy: augmenting a distributed-circuit model absorbs adversarial patterns without disrupting any single critical clean-accuracy pathway. This reversal is a core finding: optimal strategy depends on whether the classifier uses a concentrated bottleneck (Jigsaw $\rightarrow$ head suppression) or a distributed circuit (ToxiGen $\rightarrow$ augmentation).

\paragraph{\textbf{FGM training collapses catastrophically on Jigsaw.}}
FGM is stable on ToxiGen but collapses on Jigsaw after 1--3 epochs (bert\_jigsaw: 42.7\%$\rightarrow$10.9\%; roberta\_jigsaw: 69.0\%$\rightarrow$12.4\%); epoch-by-epoch curves are in Appendix~\ref{app:fgm}. The collapse is mechanistically explained: the sharp loss landscape created by L9H5/L6H1 means FGM perturbations are dominated by gradients through these bottleneck heads, causing the model to over-rely on them for clean classification and enter a brittle equilibrium that collapses within epochs. ToxiGen's distributed vulnerability profile prevents this feedback loop, yielding stable training throughout.

\subsection{\textbf{Phase 6: Split-Sample Validation}}

To confirm the identified heads are not artefacts of the specific examples used, we apply heads found on a random half (Set~A) to the held-out half (Set~B). Table~\ref{tab:generalization} shows that all four runs generalise within $\leq$1 pp, confirming the heads are stable, model-level properties rather than overfits to specific adversarial examples. Cumulative results at $k$=1,2,3,5 are in Appendix~\ref{app:split}.

\begin{table}[tb]
\centering\small
\caption{Split-sample validation: adversarial accuracy at $k$=1 on full set vs.\ held-out Set~B. All differences $\leq$1 pp.}
\label{tab:generalization}
\resizebox{\columnwidth}{!}{%
\begin{tabular}{lrrr}
\toprule
\textbf{Run} & \textbf{Full (k=1)} & \textbf{Held-out B} & \textbf{Diff} \\
\midrule
bert\_jigsaw     & 50.54\% & 50.90\% & +0.36 pp \\
roberta\_jigsaw  & 70.64\% & 71.61\% & +0.97 pp \\
bert\_toxigen    & 23.79\% & 23.10\% & $-$0.69 pp \\
roberta\_toxigen & 23.62\% & 24.02\% & +0.40 pp \\
\bottomrule
\end{tabular}}
\end{table}

\subsection{\textbf{Phase 7: Class Imbalance Sweep}}

To mechanistically confirm that the bottleneck heads act as \textit{selective toxic-class detectors}, we sweep the test-set toxic ratio from 10\%--90\% and measure the clean cost and adversarial benefit of suppression at each ratio. The clean cost scales monotonically with toxic proportion for Jigsaw (roberta\_jigsaw: +1.67 pp at 10\% toxic to +36.88 pp at 90\%, a \textbf{22$\times$ spread}) and BERT ToxiGen, confirming selective detection. RoBERTa ToxiGen inverts: L6H1 encodes a non-toxic bias, explaining its weaker head-suppression benefit in Phase~5. Full sweep data and Figure~\ref{fig:imbalance_sweep} are in Appendix~\ref{app:sweep}.

\subsection{\textbf{Extension to Llama Guard 2}}

We extend the patching pipeline to \textbf{Meta-Llama-Guard-2-8B} (32 layers $\times$ 32 heads, 1,024 total) on ToxiGen (69.2\% baseline accuracy; Jigsaw is excluded due to taxonomy mismatch). Direct PGD achieves \textbf{80.5\%} attack success; cross-model transfer from BERT/RoBERTa is negligible ($\leq$15\%). Clean activation patching identifies \textbf{L13H18} as the top crucial head ($\Delta$Loss=+0.191, $-$4.0 pp), clustered in layers 7--14. Adversarial patching identifies \textbf{L1H29} as the dominant vulnerable head ($\Delta$Loss=$-$2.759, \textbf{+37.3 pp} recovery at $k$=1); four of the top-5 vulnerable heads lie in layers 0--1. Crucially, clean and adversarial circuits are \textbf{fully decoupled}: no overlap between the top-10 clean (layers 7--14) and adversarial (layers 0--1) heads, contrasting with BERT/RoBERTa Jigsaw where both circuits coincide. This deepens the pattern: the adversarial bottleneck sits at the 75th-percentile depth for BERT (L9/12), 50th for RoBERTa (L6/12), and \textbf{3rd percentile} for Llama Guard~2 (L1/32), suggesting the exploitable circuit migrates toward the input as models scale. Full experimental detail, heatmaps (Figure~\ref{fig:llg_heatmaps}), and the bottleneck depth chart (Figure~\ref{fig:bottleneck_depth}) are in Appendix~\ref{app:llg}.

\section{\textbf{Conclusion}}

We present the first mechanistic interpretability study of toxicity classifiers, spanning BERT and RoBERTa on Jigsaw and ToxiGen, extended to Llama Guard~2. Four findings emerge: \textbf{(1)} A single attention head dominates adversarial vulnerability on Jigsaw: zeroing L9H5 (BERT) or L6H1 (RoBERTa) recovers 50.5 and 70.4 pp of adversarial accuracy at $\leq$0.6 pp clean cost, confirmed out-of-sample within $\leq$1 pp. \textbf{(2)} The optimal defence reverses by dataset: head suppression matches adversarial training on the single-bottleneck Jigsaw regime; augmentation training dominates (+45--59 pp) on the distributed ToxiGen regime; FGM collapses on Jigsaw. \textbf{(3)} The adversarial bottleneck migrates toward earlier layers as model depth grows (BERT: L9/12; RoBERTa: L6/12; Llama Guard~2: L1/32), with clean and adversarial circuits fully decoupled in Llama Guard~2. \textbf{(4)} Adversarial vulnerability is structurally unequal across the 20 Jigsaw and 13 ToxiGen demographic groups, ranging from 9.8 pp (Chinese, ToxiGen) to 100 pp (Intellectual/Learning Disability, Jigsaw) suppression recovery, exposing mechanistically traceable fairness gaps that provide principled intervention targets. 
Taken together, these results establish that the bottleneck structure of a classifier, not its overall accuracy, determines both the optimal defence strategy and the distributional fairness of that defence.

\section{\textbf{Limitations and Future Work}}

Experiments are English-only; circuit bottlenecks may differ across languages. The BERT MLM used for PGD generation may underestimate RoBERTa vulnerability. Several Jigsaw demographic groups have $n{<}30$ adversarial examples, so those per-group figures are illustrative. The Llama Guard~2 analysis does not yet measure the clean accuracy cost of head suppression. Future work should pursue: (1) online inference-time detection by monitoring vulnerable head activations; (2) automated multi-head circuit discovery \cite{conmy2023automatedcircuitdiscoverymechanistic} to improve ToxiGen robustness beyond $k$=1; (3) demographic patching for Llama Guard~2; and (4) multilingual extension to low-resource settings.

\bibliography{aaai23}

\appendix

\section{\textbf{Supplementary Figures}}
\label{app:figs}

\begin{figure}[htb]
    \centering
    \includegraphics[width=\linewidth]{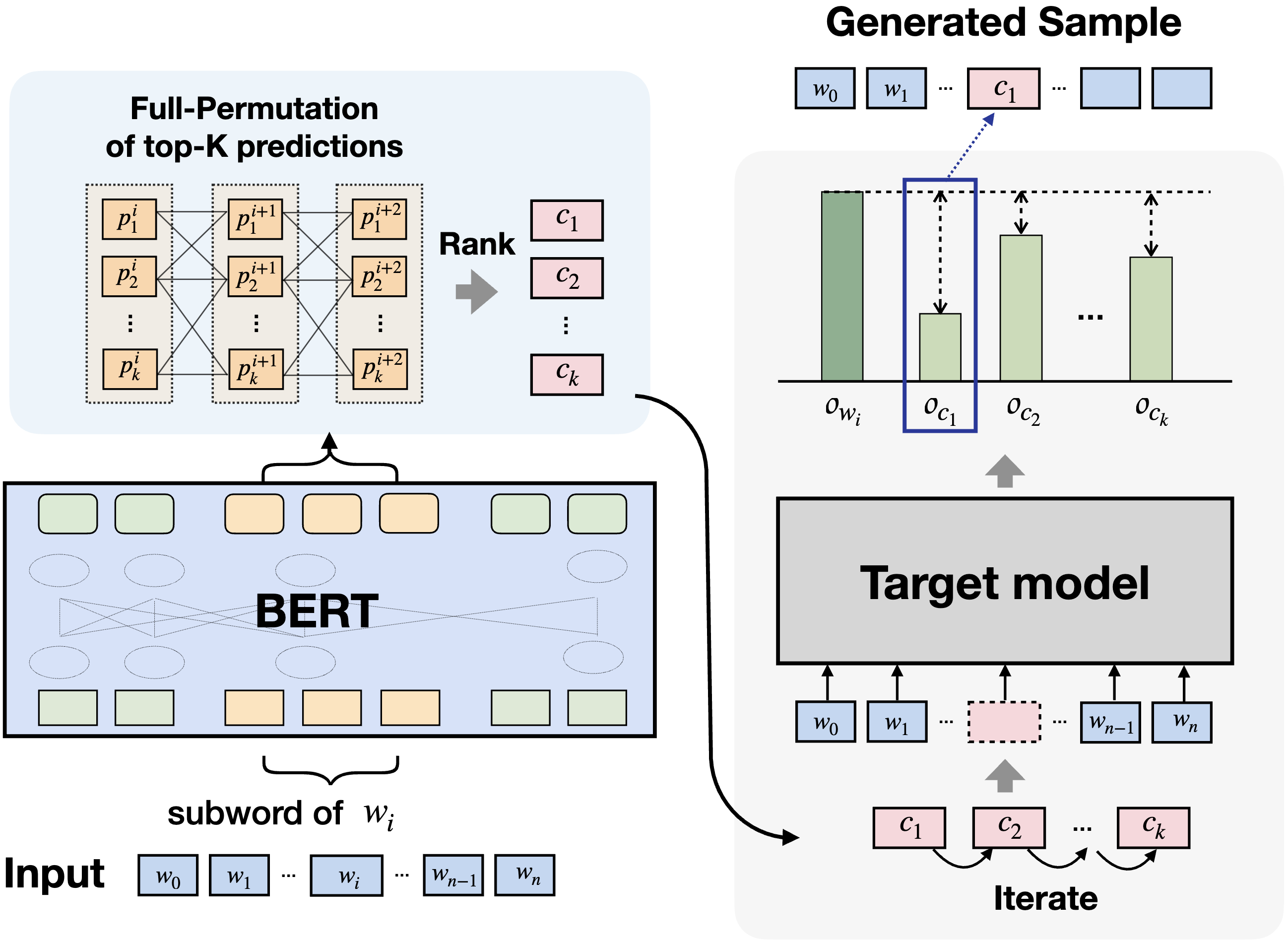}
    \caption{PGD BERT-Attack pipeline: gradients are computed w.r.t.\ input token embeddings and used to guide substitution via a BERT MLM, iterating within a norm-bounded perturbation ball.}
    \label{fig:bert_attack_diagram}
\end{figure}

\begin{figure*}[htb]
    \centering
    \includegraphics[width=\linewidth]{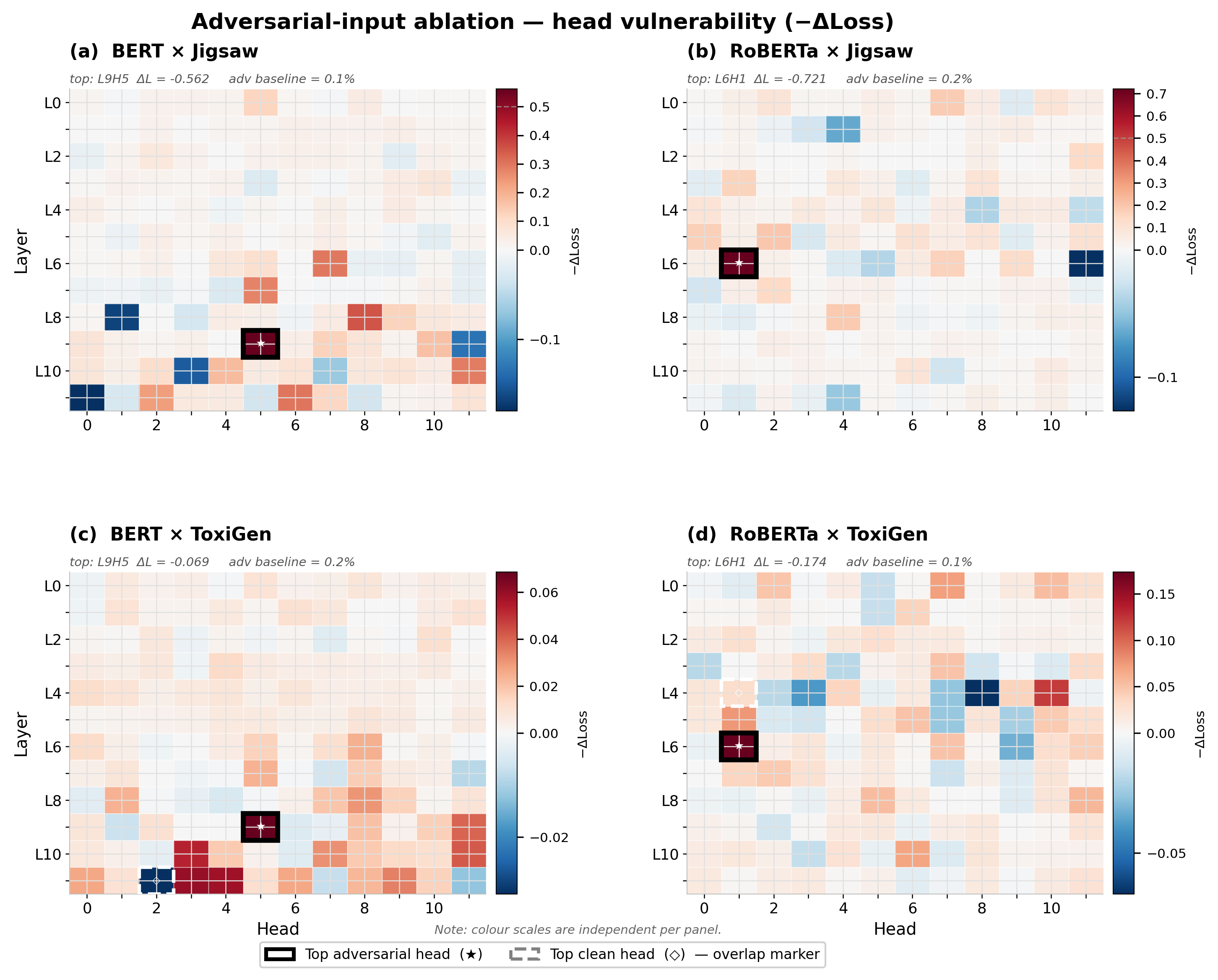}
    \caption{Adversarial-input activation patching: per-head $\Delta$Loss on successfully-attacked examples. Red cells = heads whose ablation \emph{helps} recover adversarial accuracy (negative $\Delta$Loss). L9H5 (BERT) and L6H1 (RoBERTa) are the dominant bottleneck heads for Jigsaw; ToxiGen shows a more distributed pattern.}
    \label{fig:jigsaw_heatmaps}
\end{figure*}

Figure~\ref{fig:bert_attack_diagram} illustrates the PGD BERT-Attack pipeline. Figure~\ref{fig:jigsaw_heatmaps} shows per-head $\Delta$Loss heatmaps on \emph{adversarial} inputs for all four runs. Compared with the clean-input heatmaps (Figure~\ref{fig:clean_heatmaps} in the main body), the adversarial heatmaps reveal a strong negative-$\Delta$Loss signal (heads complicit in adversarial vulnerability) that is sharply concentrated in the same bottleneck heads identified in Phase~3.

\section{\textbf{FGM Training Curves}}
\label{app:fgm}

\begin{figure}[htb]
    \centering
    \includegraphics[width=\linewidth]{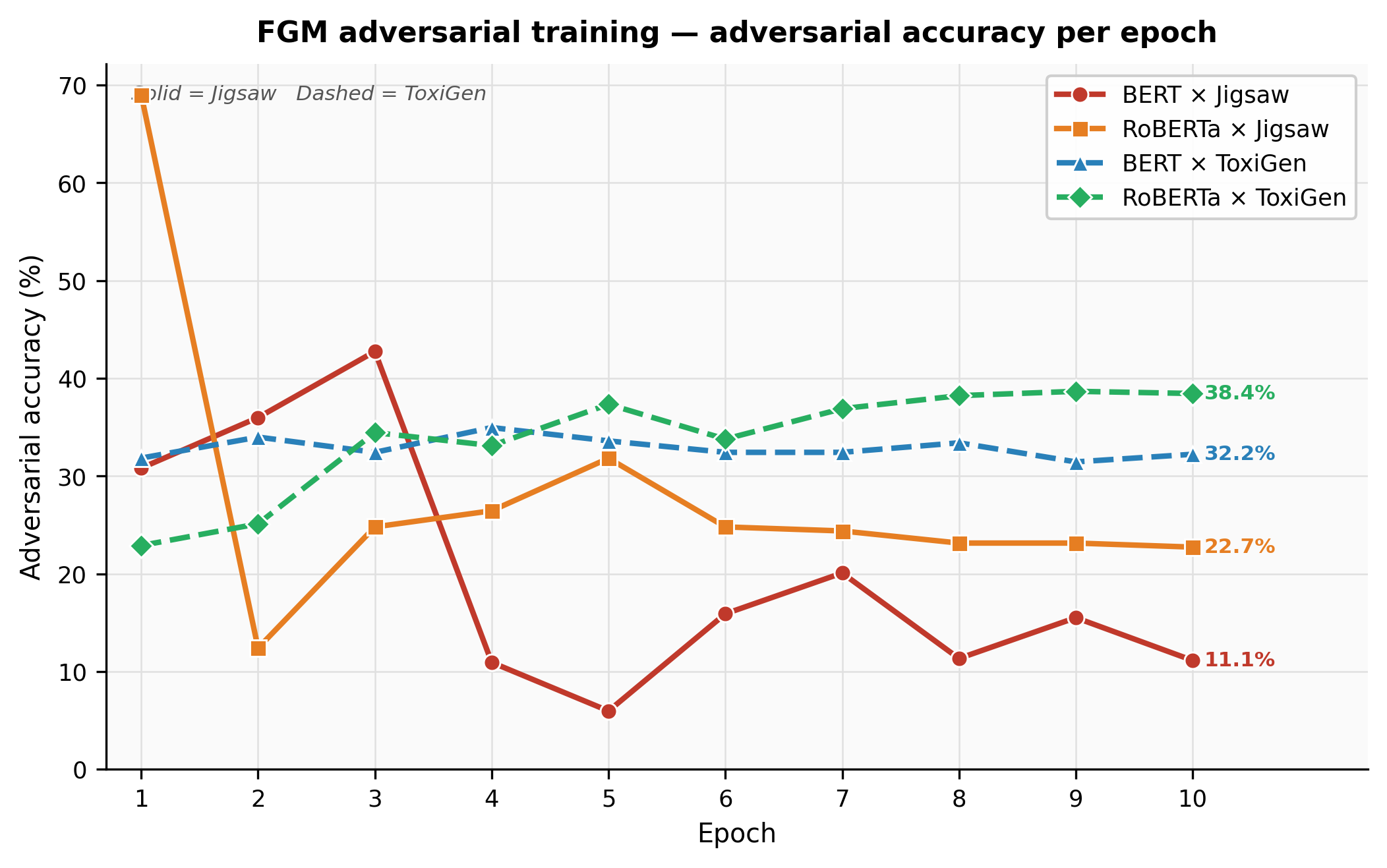}
    \caption{FGM adversarial training curves (10 epochs). Jigsaw models collapse catastrophically after the best epoch; ToxiGen models are stable throughout.}
    \label{fig:fgm_curves}
\end{figure}

Figure~\ref{fig:fgm_curves} shows epoch-by-epoch FGM adversarial training curves for all four runs. Tables~\ref{tab:fgm_jigsaw}--\ref{tab:fgm_toxigen} give the full numeric record.

\begin{table}[tb]
\centering\small
\caption{FGM training curves: BERT Jigsaw (best ep.\ 3) and RoBERTa Jigsaw (best ep.\ 1). Bold = best epoch.}
\label{tab:fgm_jigsaw}
\begin{tabular}{lrrr}
\toprule
\multicolumn{4}{c}{\textbf{BERT Jigsaw}} \\
\textbf{Ep.} & \textbf{Loss} & \textbf{Clean} & \textbf{Adv} \\
\midrule
1 & 0.122 & 94.98\% & 30.82\% \\
2 & 0.118 & 95.10\% & 35.98\% \\
\textbf{3} & \textbf{0.104} & \textbf{94.96\%} & \textbf{42.74\%} \\
4 & 0.092 & 94.63\% & 10.93\% \\
5 & 0.082 & 94.01\% & 5.96\% \\
6--10 & -- & $\sim$94\% & 11--20\% \\
\midrule
\multicolumn{4}{c}{\textbf{RoBERTa Jigsaw}} \\
\textbf{Ep.} & \textbf{Loss} & \textbf{Clean} & \textbf{Adv} \\
\midrule
\textbf{1} & \textbf{0.133} & \textbf{94.89\%} & \textbf{69.01\%} \\
2 & 0.129 & 93.46\% & 12.40\% \\
3 & 0.125 & 94.29\% & 24.79\% \\
4--10 & -- & $\sim$93--94\% & 23--32\% \\
\bottomrule
\end{tabular}
\end{table}

\begin{table}[tb]
\centering\small
\caption{FGM training curves: BERT ToxiGen (best ep.\ 4) and RoBERTa ToxiGen (best ep.\ 9). Bold = best epoch.}
\label{tab:fgm_toxigen}
\begin{tabular}{lrrr}
\toprule
\multicolumn{4}{c}{\textbf{BERT ToxiGen}} \\
\textbf{Ep.} & \textbf{Loss} & \textbf{Clean} & \textbf{Adv} \\
\midrule
1 & 0.354 & 82.57\% & 31.84\% \\
2 & 0.328 & 82.93\% & 33.98\% \\
3 & 0.276 & 83.10\% & 32.42\% \\
\textbf{4} & \textbf{0.228} & \textbf{83.16\%} & \textbf{34.96\%} \\
5 & 0.184 & 83.04\% & 33.59\% \\
6--10 & -- & 82--83\% & 31--34\% \\
\midrule
\multicolumn{4}{c}{\textbf{RoBERTa ToxiGen}} \\
\textbf{Ep.} & \textbf{Loss} & \textbf{Clean} & \textbf{Adv} \\
\midrule
1 & 0.293 & 86.62\% & 22.89\% \\
2 & 0.274 & 86.83\% & 25.11\% \\
3 & 0.245 & 86.20\% & 34.44\% \\
4 & 0.206 & 87.20\% & 33.11\% \\
5 & 0.176 & 87.08\% & 37.33\% \\
6 & 0.146 & 87.26\% & 33.78\% \\
7 & 0.118 & 87.07\% & 36.89\% \\
8 & 0.101 & 86.94\% & 38.22\% \\
\textbf{9} & \textbf{0.087} & \textbf{86.90\%} & \textbf{38.67\%} \\
10 & 0.077 & 86.98\% & 38.44\% \\
\bottomrule
\end{tabular}
\end{table}

\noindent\textbf{FGM failure analysis.} FGM collapses on Jigsaw after 1--3 epochs (BERT: epoch~3$\rightarrow$4 drop from 42.7\% to 10.9\%; RoBERTa: collapses immediately after epoch~1). The high-$\Delta$Loss bottleneck heads (L9H5/L6H1) create a sharp loss landscape: FGM perturbations are dominated by gradients from these heads, causing the model to over-rely on them and enter a brittle equilibrium. ToxiGen is far more stable (BERT: gradual decay from epoch 4; RoBERTa: monotonically improving to epoch 9), consistent with ToxiGen's more distributed vulnerability profile.

\section{\textbf{Phase 6: Split-Sample Validation (Full Results)}}
\label{app:split}

The compact k=1 comparison is in the main paper (Table~\ref{tab:generalization}). Full cumulative results for Set~B at k=1,2,3,5:

\begin{table}[tb]
\centering\small
\caption{Set B held-out cumulative adversarial accuracy (exp06).}
\label{tab:generalization_full}
\resizebox{\columnwidth}{!}{%
\begin{tabular}{llrr}
\toprule
\textbf{Run} & \textbf{k} & \textbf{Heads (Set A)} & \textbf{Held-out Adv Acc} \\
\midrule
bert\_jigsaw   & 1 & L9H5                     & 50.90\% \\
               & 2 & +L8H8                    & 65.74\% \\
               & 3 & +L11H6                   & 76.60\% \\
               & 5 & +L6H7, L10H11            & 84.68\% \\
\midrule
roberta\_jigsaw & 1 & L6H1                   & 71.61\% \\
               & 2 & +L5H2                    & 75.08\% \\
               & 3 & +L8H4                    & 76.90\% \\
               & 5 & +L0H7, L5H0              & 76.99\% \\
\midrule
bert\_toxigen  & 1 & L9H5                     & 23.10\% \\
               & 2 & +L11H3                   & 23.96\% \\
               & 3 & +L11H4                   & 24.51\% \\
               & 5 & +L10H3, L10H11           & 33.42\% \\
\midrule
roberta\_toxigen & 1 & L6H1                  & 24.02\% \\
               & 2 & +L4H10                   & 31.98\% \\
               & 3 & +L5H1                    & 35.99\% \\
               & 5 & +L10H6, L0H7             & 39.72\% \\
\bottomrule
\end{tabular}}
\end{table}

\noindent Set sizes: bert\_jigsaw A/B = 2,513/2,513; roberta\_jigsaw = 1,207/1,208; bert\_toxigen = 2,558/2,558; roberta\_toxigen = 2,248/2,248. Baseline adversarial accuracy on Set~B across all runs: 0.08--0.16\%.

\section{\textbf{Phase 7: Class Imbalance Sweep (Full Results)}}
\label{app:sweep}

\noindent\textbf{Motivation.} Phase~2 (balanced 50/50 test) shows L9H5/L6H1 ablation costing 5--12 pp clean accuracy; Phase~5 (natural Jigsaw distribution, 7.9\% toxic) shows only $-$0.2 to $-$0.6 pp. This apparent contradiction is resolved by the \emph{selective toxic-class detector} hypothesis: these heads encode toxic-positive predictions almost exclusively. If true, the clean-accuracy cost of suppression should rise monotonically with the toxic proportion of the evaluation set.

\begin{figure*}[htb]
    \centering
    \includegraphics[width=\linewidth]{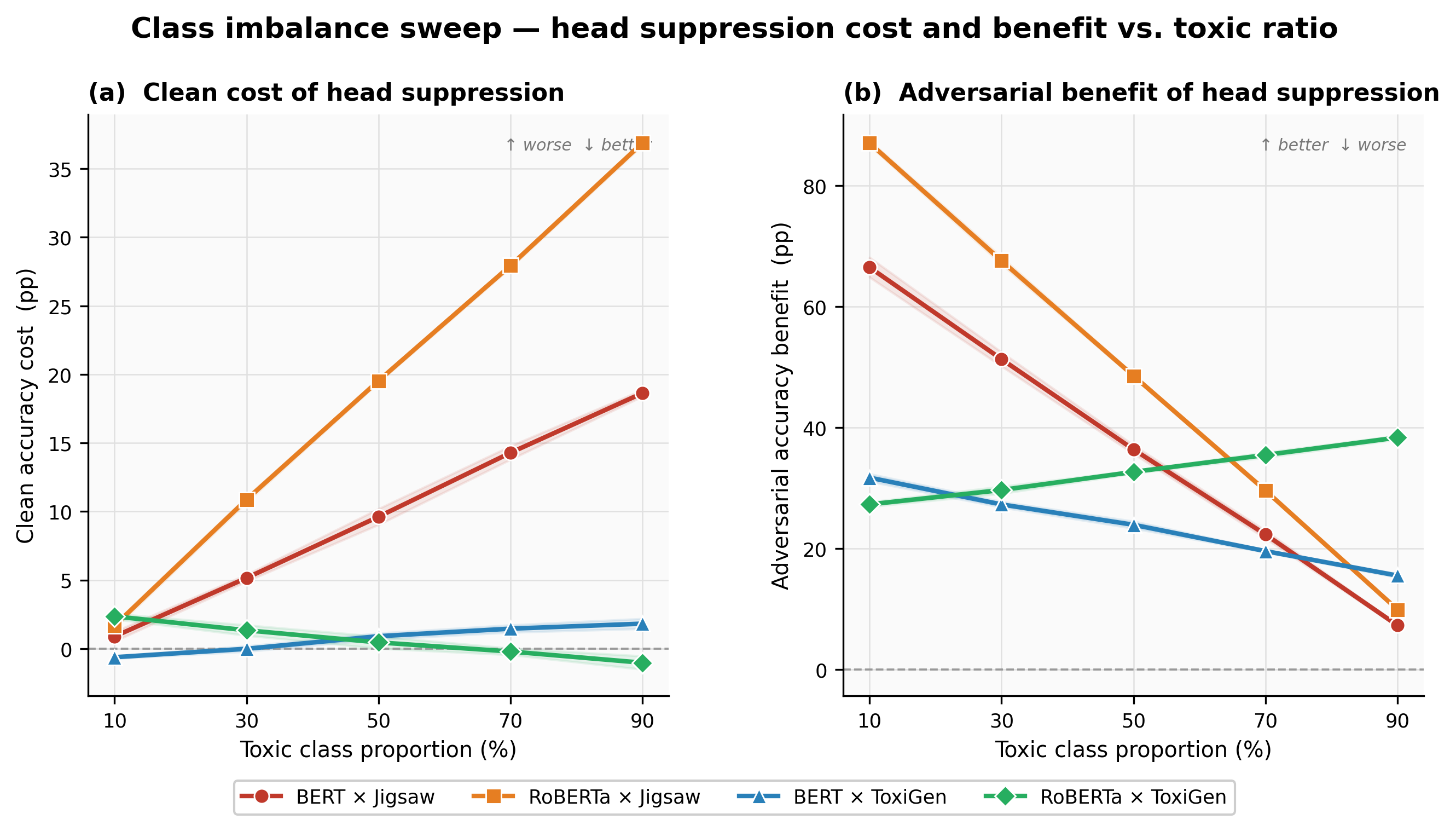}
    \caption{Class-imbalance sweep: clean cost (left) and adversarial benefit (right) of top-2 suppression at five toxic ratios. Jigsaw clean cost rises monotonically (22$\times$ spread for RoBERTa), confirming selective toxic-class detection.}
    \label{fig:imbalance_sweep}
\end{figure*}

\noindent\textbf{Protocol.} Test sets are constructed at five toxic ratios (10\%, 30\%, 50\%, 70\%, 90\%). Clean eval: $n$=2,000 per ratio (3 seeds). Adversarial eval: size-capped to $n_\text{adv}{\approx}600$--2,000 per ratio. Top-2 vulnerable heads from exp03 are suppressed. Results are shown in Figure~\ref{fig:imbalance_sweep} and Tables~\ref{tab:sweep_clean}--\ref{tab:sweep_adv}.

\begin{table}[ht]
\centering\small
\caption{Clean accuracy cost ($\Delta_\text{clean}$, pp) of top-2 head suppression at each toxic ratio.}
\label{tab:sweep_clean}
\resizebox{\columnwidth}{!}{%
\begin{tabular}{lrrrrr}
\toprule
\textbf{Run} & \textbf{10\%} & \textbf{30\%} & \textbf{50\%} & \textbf{70\%} & \textbf{90\%} \\
\midrule
bert\_jigsaw     & +0.88 & +5.15  & +9.62  & +14.28 & +18.63 \\
roberta\_jigsaw  & +1.67 & +10.87 & +19.53 & +27.93 & +36.88 \\
bert\_toxigen    & $-$0.63 & $-$0.02  & +0.90  & +1.45  & +1.82  \\
roberta\_toxigen & +2.35 & +1.33  & +0.45  & $-$0.22  & $-$1.03  \\
\bottomrule
\end{tabular}}
\end{table}

\begin{table}[ht]
\centering\small
\caption{Adversarial accuracy benefit ($\Delta_\text{adv}$, pp) of top-2 head suppression at each toxic ratio.}
\label{tab:sweep_adv}
\resizebox{\columnwidth}{!}{%
\begin{tabular}{lrrrrr}
\toprule
\textbf{Run} & \textbf{10\%} & \textbf{30\%} & \textbf{50\%} & \textbf{70\%} & \textbf{90\%} \\
\midrule
bert\_jigsaw     & +66.51 & +51.31 & +36.38 & +22.31 & +7.28  \\
roberta\_jigsaw  & +87.06 & +67.59 & +48.56 & +29.58 & +9.90  \\
bert\_toxigen    & +31.72 & +27.28 & +23.92 & +19.57 & +15.52 \\
roberta\_toxigen & +27.30 & +29.67 & +32.67 & +35.47 & +38.35 \\
\bottomrule
\end{tabular}}
\end{table}

\noindent\textbf{Interpretation.} Three of four runs pass the clean-cost monotonicity test, directly confirming the selective-detector hypothesis. The exception, roberta\_toxigen (inverted direction), is informative: L6H1 in that run appears to encode \emph{non-toxic} prediction bias; suppression nudges predictions toward toxic, which \emph{helps} on majority-toxic sets. This is consistent with L6H1's weaker head-suppression effect on ToxiGen adversarial accuracy (only +29.8 pp vs +70.4 pp for Jigsaw). Adversarial benefit decreases with toxic ratio for Jigsaw/BERT-ToxiGen because the PGD pool is imbalanced (7.9\% and 11\% toxic for Jigsaw/BERT; at high toxic ratios, fewer non-toxic adversarial examples remain to be recovered). RoBERTa ToxiGen's adv benefit increases because its near-balanced PGD pool ($\sim$45\% toxic) means more toxic examples benefit from suppression as the ratio rises.

\section{\textbf{Llama Guard 2: Full Experimental Detail}}
\label{app:llg}

\subsection*{\textbf{A. Exp08: Transfer Attack Evaluation}}

PGD adversarial examples crafted against each of the four BERT/RoBERTa models were passed through Llama Guard~2 to measure cross-model transfer. Transfer rate = fraction of PGD-successful examples where Llama Guard was correct on the original text but misclassifies the adversarial version.

\begin{table}[ht]
\centering\small
\caption{Transfer attack results: BERT/RoBERTa adversaries vs.\ Llama Guard~2.}
\label{tab:llg_transfer}
\resizebox{\columnwidth}{!}{%
\begin{tabular}{lrrrr}
\toprule
\textbf{Source} & \textbf{Orig} & \textbf{Adv} & \textbf{Transfer} & \textbf{Toxic cls} \\
\midrule
bert\_jigsaw     & 90.5\% & 88.3\% & 3.1\%  & 26.3\% \\
roberta\_jigsaw  & 90.5\% & 90.0\% & 1.8\%  & 18.9\% \\
bert\_toxigen    & 68.5\% & 60.6\% & 14.6\% & 42.2\% \\
roberta\_toxigen & 68.5\% & 63.6\% & 8.3\%  & 30.8\% \\
\bottomrule
\end{tabular}}
\end{table}

Jigsaw transfer is negligible (1.8--3.1\%) because Llama Guard's safety taxonomy (S1--S11) does not cover Jigsaw-style rude/offensive speech; Llama Guard achieves only $\sim$6.5\% toxic-class accuracy on Jigsaw. ToxiGen transfer is partial but non-trivial (8--15\% overall; 31--42\% on the toxic class), reflecting genuine alignment between ToxiGen's demographic hate speech and Llama Guard's S9 (Hate) category. Transfer is dataset-dependent, not architecture-dependent.

\subsection*{\textbf{B. Exp09: Direct PGD on Llama Guard 2 / ToxiGen}}

Using the same PGD loop adapted to Llama Guard's first-token logits (user-text embeddings only; template tokens frozen; 6 GPUs, $n$=2,000 stratified ToxiGen examples), direct attack achieves \textbf{80.45\%} success (1,609/2,000 examples; per-worker range 76.6--83.2\%). This confirms that Llama Guard~2 is highly vulnerable to gradient-guided attacks when the adversary has model access, despite its 8B-parameter safety-oriented design. The contrast with near-zero transfer from BERT/RoBERTa attacks confirms that vulnerability is real but model-specific.

\subsection*{\textbf{C. Exp10: Clean Attention Patching (Llama Guard 2)}}

Zero-ablation of each of 1,024 heads (32 layers $\times$ 32 heads) on 500 balanced ToxiGen examples. Baseline accuracy: 69.2\%. Top-10 crucial heads:

\begin{table}[ht]
\centering\small
\caption{Llama Guard~2: top-10 crucial heads on clean ToxiGen input (exp10).}
\label{tab:llg_clean_heads}
\begin{tabular}{rllrr}
\toprule
\textbf{Rank} & \textbf{Layer} & \textbf{Head} & \textbf{$\Delta$Loss} & \textbf{Acc drop} \\
\midrule
1  & 13 & 18 & +0.191 & $-$4.00 pp \\
2  & 7  & 29 & +0.151 & $-$2.40 pp \\
3  & 9  & 0  & +0.114 & $-$0.80 pp \\
4  & 12 & 2  & +0.109 & $-$1.40 pp \\
5  & 11 & 25 & +0.104 & $-$0.80 pp \\
6  & 12 & 20 & +0.089 & +0.40 pp \\
7  & 12 & 6  & +0.086 & $-$0.80 pp \\
8  & 8  & 19 & +0.081 & $-$1.60 pp \\
9  & 14 & 8  & +0.081 & 0.00 pp  \\
10 & 11 & 24 & +0.080 & $-$0.20 pp \\
\bottomrule
\end{tabular}
\end{table}

Clean crucial heads cluster in layers 7--14 (mid-depth of the 32-layer network), consistent with mid-level semantic processing of hate speech content. L13H18 is the dominant clean head ($\Delta$Loss $= +0.191$), causing a 4.0 pp accuracy drop, larger than any BERT/RoBERTa head in absolute $\Delta$Loss terms, but smaller in relative accuracy impact, consistent with a more distributed processing architecture.

\subsection*{\textbf{D. Exp11: Adversarial Attention Patching (Llama Guard 2)}}

Zero-ablation on 1,000 PGD-successful adversarial examples. Baseline adversarial accuracy: 3.3\%. Top-10 vulnerable heads and cumulative suppression results:

\begin{table}[ht]
\centering\small
\caption{Llama Guard~2: top-10 vulnerable heads on adversarial ToxiGen input (exp11).}
\label{tab:llg_adv_heads}
\begin{tabular}{rllrr}
\toprule
\textbf{Rank} & \textbf{Layer} & \textbf{Head} & \textbf{$\Delta$Loss} & \textbf{Adv acc $\Delta$} \\
\midrule
1  & 1  & 29 & $-$2.759 & +37.3 pp \\
2  & 0  & 29 & $-$0.565 & +33.0 pp \\
3  & 1  & 28 & $-$0.387 & +20.4 pp \\
4  & 0  & 2  & $-$0.378 & +25.5 pp \\
5  & 12 & 22 & $-$0.376 & +28.8 pp \\
6  & 13 & 31 & $-$0.337 & +5.8 pp  \\
7  & 0  & 31 & $-$0.310 & +27.5 pp \\
8  & 12 & 0  & $-$0.237 & +3.6 pp  \\
9  & 14 & 24 & $-$0.234 & +7.0 pp  \\
10 & 31 & 14 & $-$0.220 & $-$0.3 pp  \\
\bottomrule
\end{tabular}
\end{table}

\begin{table}[ht]
\centering\small
\caption{Llama Guard~2: cumulative head suppression adversarial accuracy (exp11).}
\label{tab:llg_suppression}
\begin{tabular}{rlr}
\toprule
\textbf{k} & \textbf{Heads suppressed} & \textbf{Adv acc} \\
\midrule
0 & -- & 3.3\% \\
1 & L1H29 & 40.6\% \\
2 & +L0H29 & 41.3\% \\
3 & +L1H28 & 38.2\% \\
5 & +L0H2, L12H22 & 43.5\% \\
\bottomrule
\end{tabular}
\end{table}

Four of the top-5 vulnerable heads sit in layers 0--1, a striking contrast to BERT (L9/12) and RoBERTa (L6/12). The clean-circuit heads (layers 7--14) and adversarial-circuit heads (layers 0--1) are \emph{fully decoupled}: zero overlap between the top-10 clean and top-10 adversarial head lists. This is the only run where clean and adversarial circuits are completely disjoint.

\begin{figure*}[htb]
    \centering
    \includegraphics[width=\linewidth]{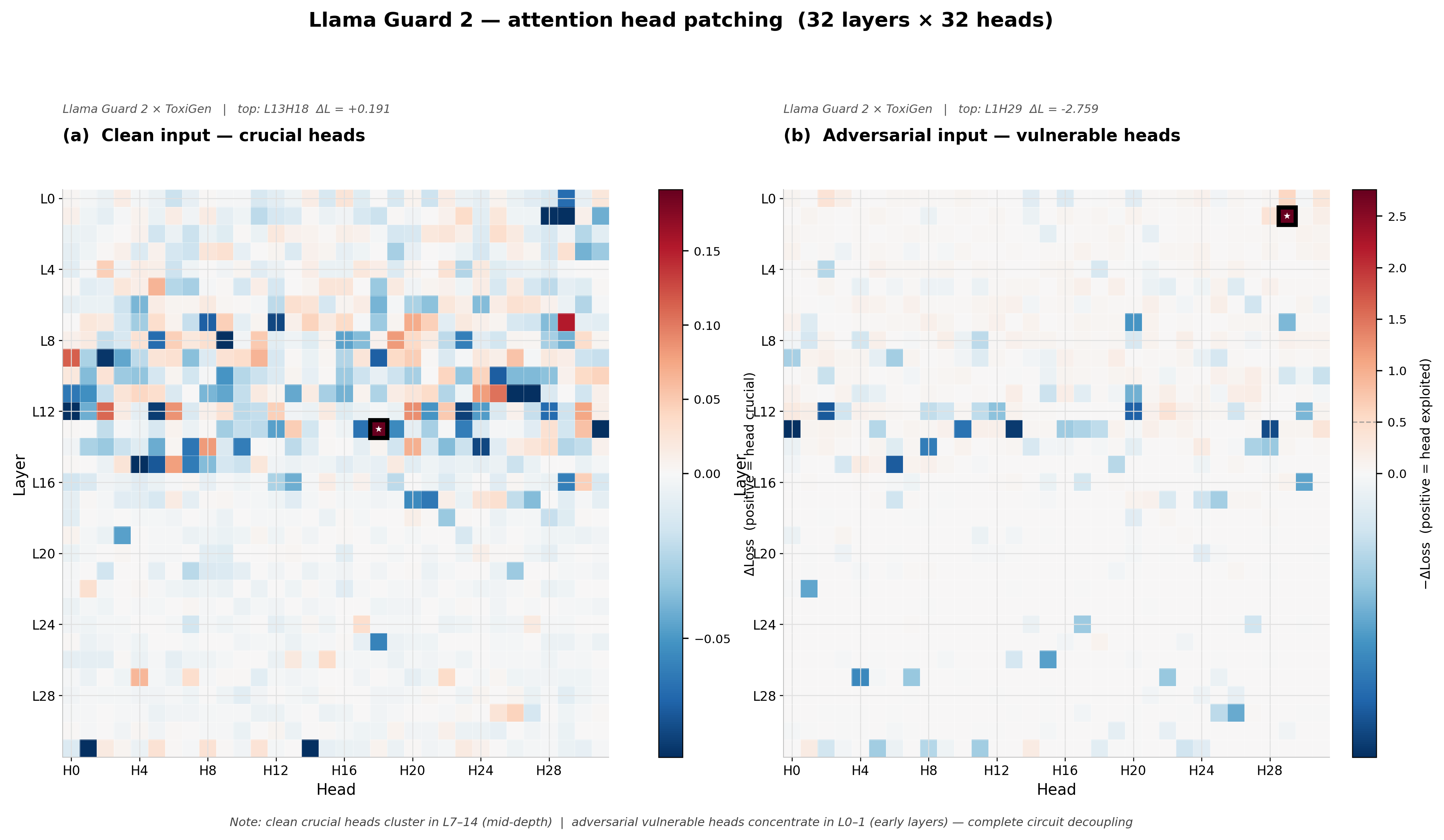}
    \caption{Llama Guard~2 activation patching heatmaps (32 layers $\times$ 32 heads). Left: clean-input $\Delta$Loss; right: adversarial-input $\Delta$Loss. The clean circuit concentrates in layers 7--14; the adversarial circuit concentrates in layers 0--1. The two circuits are fully decoupled.}
    \label{fig:llg_heatmaps}
\end{figure*}

\begin{figure}[htb]
    \centering
    \includegraphics[width=\linewidth]{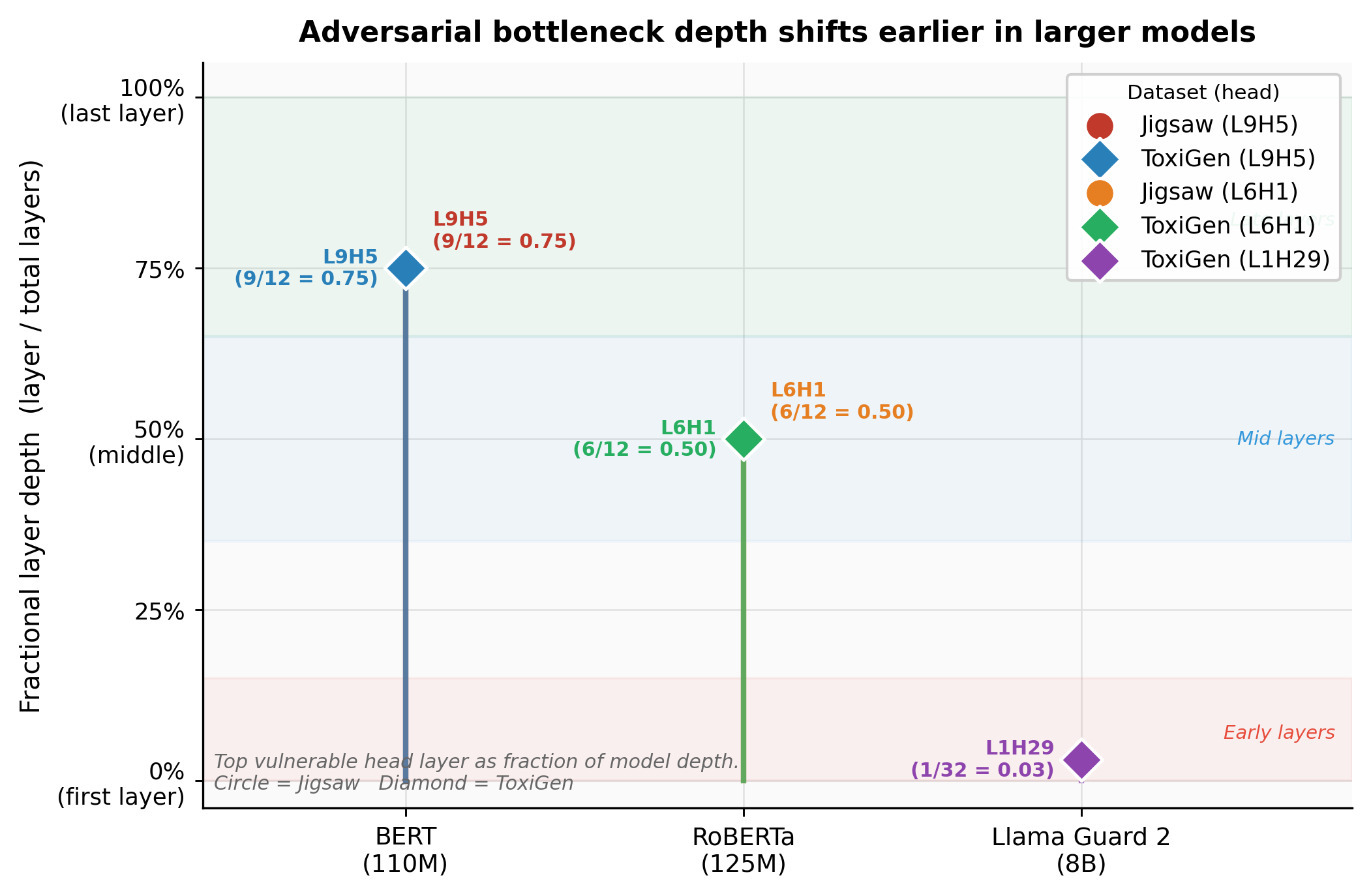}
    \caption{Bottleneck layer depth (as a percentile of model depth) for the top vulnerable head across all five model runs. BERT: 75th percentile (L9/12); RoBERTa: 50th percentile (L6/12); Llama Guard~2: 3rd percentile (L1/32). The bottleneck migrates to earlier layers in larger, decoder-based models.}
    \label{fig:bottleneck_depth}
\end{figure}

\subsection*{\textbf{E. Circuit Decoupling and Architectural Comparison}}

The adversarial bottleneck migrates to progressively earlier layers as model depth increases: L9/12 (75th percentile, BERT) $\to$ L6/12 (50th percentile, RoBERTa) $\to$ L1/32 (3rd percentile, Llama Guard~2). A plausible explanation is that larger, decoder-based models develop high-level semantic representations earlier in the network, so the surface-feature exploit that PGD relies on is concentrated at the shallowest layer of non-embedding computation. The complete clean/adversarial circuit decoupling in Llama Guard~2 ToxiGen (absent in all BERT/RoBERTa runs and in the Llama Guard~2 Jigsaw run) suggests that on ToxiGen, the 8B model has learned to use entirely separate representations for clean hate-speech detection vs.\ the features that PGD exploits.

\end{document}